\documentclass[10pt,journal,compsoc]{IEEEtran}
\ifCLASSOPTIONcompsoc
  \usepackage[nocompress]{cite}
\else
  \usepackage{cite}
\fi
\ifCLASSINFOpdf
   \usepackage[pdftex]{graphicx}
\else
\fi
\usepackage{amsmath,amssymb,amsfonts}
\usepackage[linesnumbered,ruled,vlined]{algorithm2e}

\usepackage{array}

\ifCLASSOPTIONcompsoc
 \usepackage[caption=false,font=footnotesize,labelfont=sf,textfont=sf]{subfig}
\else
 \usepackage[caption=false,font=footnotesize]{subfig}
\fi

\usepackage{stfloats}
\usepackage{url}

\usepackage{booktabs}
\usepackage{breqn}
\usepackage{footnote}

\makesavenoteenv{tabular}
\usepackage{multirow}
\makesavenoteenv{table}

\usepackage{balance}
\usepackage{placeins}
\usepackage{bbm}
\usepackage{scalerel}
\usepackage{tikz}
\usetikzlibrary{svg.path}

\definecolor{orcidlogocol}{HTML}{A6CE39}
\tikzset{
  orcidlogo/.pic={
    \fill[orcidlogocol] svg{M256,128c0,70.7-57.3,128-128,128C57.3,256,0,198.7,0,128C0,57.3,57.3,0,128,0C198.7,0,256,57.3,256,128z};
    \fill[white] svg{M86.3,186.2H70.9V79.1h15.4v48.4V186.2z}
                 svg{M108.9,79.1h41.6c39.6,0,57,28.3,57,53.6c0,27.5-21.5,53.6-56.8,53.6h-41.8V79.1z M124.3,172.4h24.5c34.9,0,42.9-26.5,42.9-39.7c0-21.5-13.7-39.7-43.7-39.7h-23.7V172.4z}
                 svg{M88.7,56.8c0,5.5-4.5,10.1-10.1,10.1c-5.6,0-10.1-4.6-10.1-10.1c0-5.6,4.5-10.1,10.1-10.1C84.2,46.7,88.7,51.3,88.7,56.8z};
  }
}

\newcommand\orcidicon[1]{\href{https://orcid.org/#1}{\mbox{\scalerel*{
\begin{tikzpicture}[yscale=-1,transform shape]
\pic{orcidlogo};
\end{tikzpicture}
}{|}}}}
\usepackage{hyperref}
\usepackage{paralist}
\usepackage{textcomp}
\usepackage{xcolor}
\usepackage{cleveref}

\SetCommentSty{mycommfont}

\SetKwInput{KwInput}{Input}                
\SetKwInput{KwOutput}{Output}              

\def\BibTeX{{\rm B\kern-.05em{\sc i\kern-.025em b}\kern-.08em
 T\kern-.1667em\lower.7ex\hbox{E}\kern-.125emX}}
 
\DeclareMathOperator*{\argmax}{arg\,max}

\newcommand{\norm}[1]{\left\lVert#1\right\rVert}
\newcommand{\fatx}{\mathbf{x}}

\newcommand{\faty}{\mathbf{y}}

\newcommand{\rulesep}{\unskip\ \vrule\ }

\begin{document}

\definecolor{mypink1}{rgb}{0.858, 0.188, 0.478}
\definecolor{mypink2}{RGB}{219, 48, 122}

\newcommand{\smallmethodname}[1]{{CUDA}}
\newcommand{\fullmethodname}[1]{Contradistinguisher for Unsupervised Domain Adaptation (\smallmethodname{})}
\newcommand{\fullmethodnametitle}[1]{{Contradistinguisher: A Vapnik's Imperative to Unsupervised Domain Adaptation}}


\title{\fullmethodnametitle{}}
\author{Sourabh~Balgi~\orcidicon{0000-0002-3329-5533} and
        Ambedkar~Dukkipati~\orcidicon{0000-0002-6352-6283}
        \IEEEcompsocitemizethanks{\IEEEcompsocthanksitem S. Balgi and A. Dukkipati (Contact Author) are with the Department
of Computer Science and Automation, Indian Institute of Science, Bengaluru,
Karnataka, India, 560012.\newline\protect~E-mail: \{{sourabhbalgi}, {ambedkar}\}@iisc.ac.in}\\
}

%
%

\markboth{IEEE Transactions on Pattern Analysis and Machine
  Intelligence (Accepted)}%
{Balgi \MakeLowercase{\textit{et al.}}: \fullmethodnametitle}
%



\IEEEtitleabstractindextext{%
\begin{abstract}
Recent domain adaptation works rely on an indirect way of first aligning the source and target domain distributions and then train a classifier on the labeled source domain to classify the target domain. However, the main drawback of this approach is that obtaining a near-perfect domain alignment in itself might be difficult/impossible (e.g., language domains). 
To address this, {inspired by how humans use supervised-unsupervised learning to perform tasks seamlessly across multiple domains or tasks}, we follow Vapnik's imperative of statistical learning that states any desired problem should be solved in the most direct way rather than solving a more general intermediate task and propose a direct approach to domain adaptation that does not require domain alignment. We propose a model referred to as Contradistinguisher that learns contrastive features and whose objective is to jointly learn to contradistinguish the unlabeled target domain in an unsupervised way and classify in a supervised way on the source domain. {We achieve the state-of-the-art on Office-31, Digits and VisDA-2017 datasets in both single-source and multi-source settings. We demonstrate that performing data augmentation results in an improvement in the performance over vanilla approach. We also notice that the contradistinguish-loss enhances performance by increasing the shape bias.}

\end{abstract}

}

\maketitle

\IEEEdisplaynontitleabstractindextext

%
\IEEEpeerreviewmaketitle

\section{Introduction}
\IEEEPARstart{T}{he} recent success of deep neural networks for supervised learning tasks in several areas like computer vision, speech, and natural language processing can be attributed to the models trained on large amounts of labeled data. However, acquiring massive amounts of labeled data in some domains can be very expensive or not possible at all. Additionally, the amount of time required for labeling the data to use existing deep learning techniques can be very high initially for a new domain. This is known as \emph{{cold-start}}. On the contrary, cost-effective unlabeled data can be easily obtained in large amounts for most new domains. So, one can aim to transfer the knowledge from a labeled source domain to perform tasks on an unlabeled target domain.
To study this, under the purview of transductive transfer learning, several approaches like domain adaptation, sample selection bias, co-variance shift have been explored in recent times. 

Existing domain adaptation approaches mostly rely on domain alignment, i.e., align both domains so that they are superimposed and indistinguishable in the latent space. This domain alignment can be achieved in three main ways:
\begin{inparaenum}[(a)] 
\item discrepancy-based methods~\cite{DBLP:conf/icml/LongC0J15, DBLP:conf/nips/LongZ0J16, DBLP:conf/icml/LongZ0J17, DBLP:conf/iccv/HausserFMC17, 8578490, french2018selfensembling, DBLP:journals/corr/LouizosSLWZ15, 2017arXiv170208811Z, rozantsev2018beyond, 8792192, mancini2018boosting, cariucci2017autodial, carlucci2017just, gopalan2011domain, xie2015learning,  DBLP:conf/cvpr/PanYLWNM19, DBLP:conf/cvpr/Kang0YH19, DBLP:conf/iccv/PengBXHSW19},
\item reconstruction-based methods~\cite{10.1007/978-3-319-46493-0_36, Bousmalis:2016:DSN:3157096.3157135} and
\item adversarial adaptation methods~\cite{pmlr-v37-ganin15, ganin2016domain,  NIPS2016_6544, 8099799, DBLP:conf/cvpr/Sankaranarayanan18a, DBLP:conf/cvpr/LiuYFWCW18, Russo_2018_CVPR, pmlr-v80-hoffman18a, xie2018learning, NIPS2018_7436, DBLP:conf/aaai/ChenCJJ19, shu2018a, hosseini-asl2018augmented, liang2018aggregating, xu2018deep, chen2019transferability, wang2019transferable, 9080115, DBLP:conf/cvpr/KurmiKN19, DBLP:conf/cvpr/ChenXHRD0XH19
}.
\end{inparaenum}%

These domain alignment strategies of indirectly addressing the task of unlabeled target domain classification have three main drawbacks. 
\begin{inparaenum}[(i)]
(i) The sub-task of obtaining a perfect alignment of the domain in itself might be impossible or very difficult due to large domain shifts (e.g., language domains). 
(ii) The use of multiple classifiers and/or GANs to align the distributions unnecessarily increases the complexity of the neural networks leading to over-fitting in many cases.
(iii) Due to distribution alignment, the domain-specific information is lost as the domains get morphed. 
\end{inparaenum}

A particular case where the domain alignment and the classifier trained on the source domain might fail is that the target domain is more suited to classification tasks than the source domain with lower classification performance. In this case, it is advised to perform the classification directly on the unlabeled target domain in an unsupervised manner as domain alignment onto a less suited source domain only leads to loss of information. It is reasonable to assume that for the main objective of unlabeled target domain classification, one can use all the information in the target domain and optionally incorporate any useful information from the labeled source domain and not the other way around. These drawbacks push us to challenge the idea of solving domain adaptation problems without solving the general problem of domain alignment. 

In this work, we study unsupervised domain adaptation by learning contrastive features in the unlabeled target domain in a fully unsupervised manner with the help of a classifier simultaneously trained on the labeled source domain.
{More importantly, we derive our motivation from the \textbf{Vapnik's imperative} that motivated the statistical learning theory~\cite{vapnik1999overview, vapnik2013nature}.
\begin{quote}
    \textit{``When solving a given problem, try to avoid solving a more general problem as an intermediate step.''}~\cite{vapnik2013nature}
    \end{quote}
}

In the context of domain adaptation, the desired problem is classification on the unlabeled target domain, and domain alignment followed by most standard methods is the general intermediate. 
Considering the various drawback of the domain alignment approach, in this paper, we propose a method for domain adaptation that does not require domain alignment and approach the problem directly. 



This work extends our earlier conference paper~\cite{Balgi2019CUDA} in the following way.
\begin{inparaenum}[(i)]
\item We provide additional experimental results on more complex domain adaptation dataset \href{https://drive.google.com/open?id=0B4IapRTv9pJ1WGZVd1VDMmhwdlE}{Office-31}~\cite{DBLP:conf/eccv/SaenkoKFD10} which includes images from three different sources, AMAZON~($\mathcal{A}$), DSLR~($\mathcal{D}$) and WEBCAM~($\mathcal{W}$) categorized into three domains respectively with only a few labeled high-resolution images.
{\item We provide additional experimental results on the benchmark \href{http://ai.bu.edu/visda-2017/}{VisDA-2017}~\cite{peng2017visda} dataset for unsupervised domain adaptation that includes synthetic and real-world images from 12 different classes.
\item We provide several ablation studies and demonstrations that will provide insights into the working of our proposed method~\smallmethodname{}~\cite{Balgi2019CUDA}. Additionally, we observed that the proposed contradistinguish loss helps to learn high-level features related to the shapes of the objects in the image.}
\item We extend our algorithm to the case of multi-source domain adaptation and establish benchmark results on \href{https://drive.google.com/open?id=0B4IapRTv9pJ1WGZVd1VDMmhwdlE}{Office-31}~\cite{DBLP:conf/eccv/SaenkoKFD10} dataset and \textcolor{black}{Digits datasets}.
\end{inparaenum}
\begin{figure*}[t!]
\begin{center}
\subfloat[$D0{ \rightarrow }D1$ using standard\newline domain alignment method.]{
\includegraphics[width=0.215\linewidth,height=0.215\textheight,keepaspectratio=true]{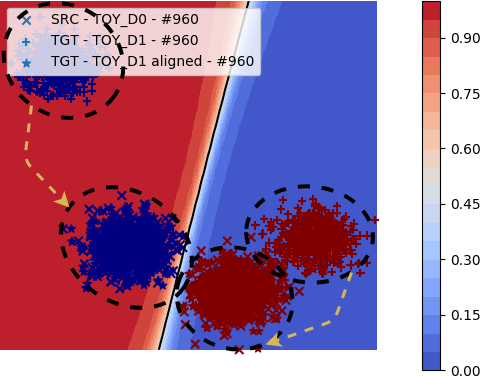}}
\hfill
\subfloat[$D1{ \rightarrow }D0$ using standard\newline domain alignment method.]{
\includegraphics[width=0.215\linewidth,height=0.215\textheight,keepaspectratio=true]{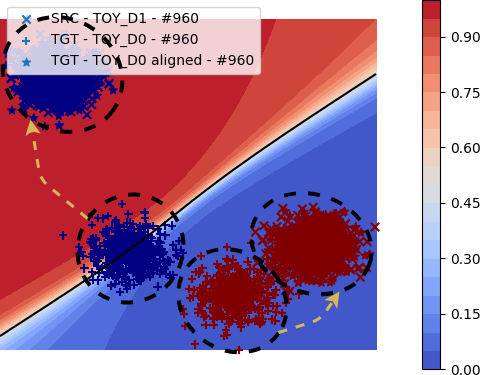}}
\hfill
\rulesep
\hfill
\rulesep
\subfloat[$D0{ \rightarrow }D1$ using standard\newline domain alignment method.]{
\includegraphics[width=0.215\linewidth,height=0.215\textheight,keepaspectratio=true]{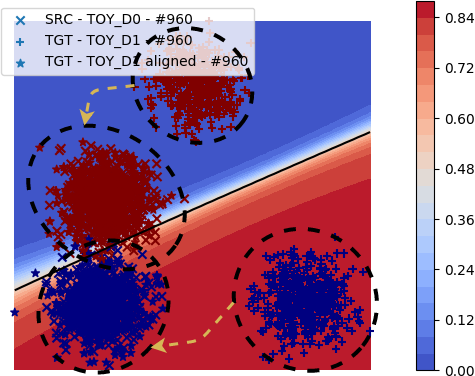}\label{sfig:cp_src_toy_d0_seed_3234_ss_contour_epoch}}
\hfill
\subfloat[$D1{ \rightarrow }D0$ using standard\newline domain alignment method.]{
\includegraphics[width=0.215\linewidth,height=0.215\textheight,keepaspectratio=true]{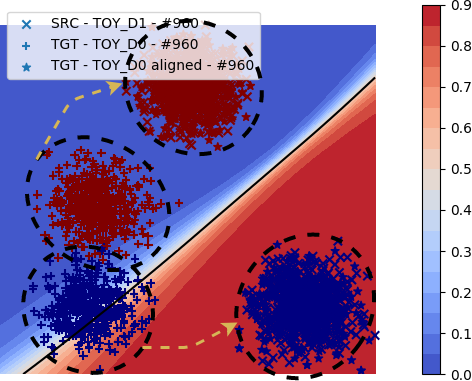}\label{sfig:cp_src_toy_d1_seed_3234_ss_contour_epoch}}
\\
\subfloat[$D0{ \rightarrow }D1$ using our\newline proposed method \smallmethodname{}.]{
\includegraphics[width=0.215\linewidth,height=0.215\textheight,keepaspectratio=true]{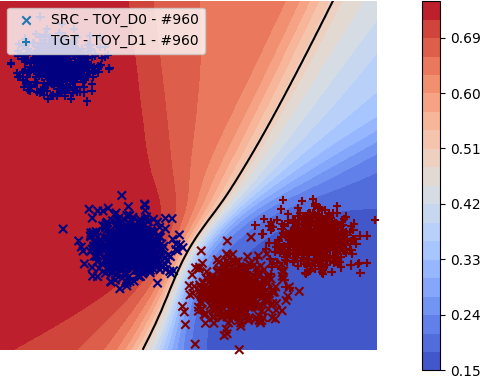}\label{sfig:cp_src_toy_d0_seed_0022_ss_tu_ta_contour_epoch}}
\hfill
\subfloat[$D1{ \rightarrow }D0$ using our\newline proposed method \smallmethodname{}.]{
\includegraphics[width=0.215\linewidth,height=0.215\textheight,keepaspectratio=true]{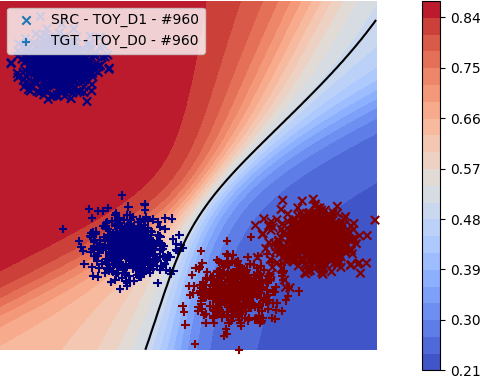}\label{sfig:cp_src_toy_d1_seed_0022_ss_tu_ta_contour_epoch}}
\hfill
\rulesep
\hfill
\rulesep
\subfloat[$D0{ \rightarrow }D1$ using our\newline proposed method \smallmethodname{}.]{
\includegraphics[width=0.215\linewidth,height=0.215\textheight,keepaspectratio=true]{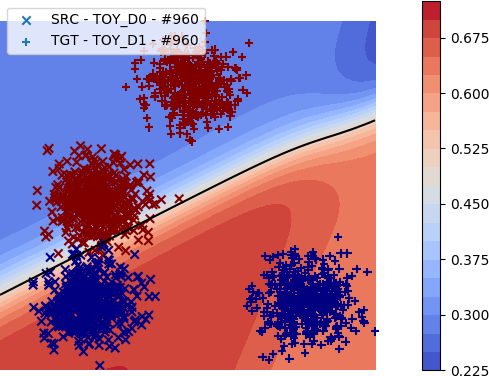}\label{sfig:cp_src_toy_d0_seed_3234_ss_tu_ta_contour_epoch}}
\hfill
\subfloat[$D1{ \rightarrow }D0$ using our\newline proposed method \smallmethodname{}.]{
\includegraphics[width=0.215\linewidth,height=0.215\textheight,keepaspectratio=true]{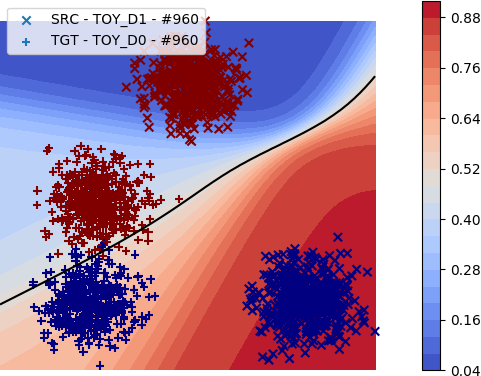}\label{sfig:cp_src_toy_d1_seed_3234_ss_tu_ta_contour_epoch}}
\caption{{Demonstration of difference in domain alignment and proposed method \smallmethodname{} on the {$2$-dimensional} blobs synthetic toy-dataset for domain distributions from popular $scikit{ - }learn$~\cite{pedregosa2011scikit}.
We provide two example settings of different data distributions for domains $D0$ and $D1$ in both the directions of domain adaptation $D0{ \leftrightarrow }D1$ separated by the vertical lines.
In each setting, the top row corresponds to the domain alignment approach. The bottom row corresponds to the proposed method \smallmethodname{} compared to their respective domain alignment in the top row in both $D0{ \leftrightarrow }D1$ domain adaptation tasks. The yellow dotted lines indicate the domain alignment process to superimpose the target domain onto the source domain, thereby morphing both the domains.
similarly, the two sub-columns indicate the experiments with swapped source and target domains. 
Unlike the domain alignment approach, where the classifier is learnt only on source domain, \smallmethodname{} demonstrates the contradistinguisher jointly learnt to classify on both the domains. 
As seen above, swapping domains affect the classifier learnt in domain alignment because the classifier depends on the source domain. However, because of joint learning on both the domains simultaneously, \emph{contradistinguisher} shows almost the same decision boundary irrespective of the source domain, i.e., irrespective of the direction of the domain adaptation, i.e., $D0{ \rightarrow }D1$ or $D1{ \rightarrow }D0$. 
(Best viewed in color.)
}}
\label{fig:toy dataset demo small}
\end{center}
\end{figure*}

A summary of our contributions in this paper is as follows.
\begin{enumerate}
 \item We propose a simple method \fullmethodname{} that directly addresses the problem of domain adaptation by learning a single classifier, which we refer to as Contradistinguisher, jointly in an unsupervised manner over the unlabeled target domain and in a supervised manner over the labeled source domain. Hence, overcoming the drawbacks of distribution alignment-based techniques. 
 \item We formulate a `contradistinguish loss' to directly utilize unlabeled target domain and address the classification task using unsupervised feature learning. Note that a similar approach called DisCoder~\cite{Pandey2017UnsupervisedFL} was used for a much simpler task of semi-supervised feature learning on a single domain with no domain distribution shift. 
{ \item We extend our experiments to more complex domain adaptation datasets \href{https://drive.google.com/open?id=0B4IapRTv9pJ1WGZVd1VDMmhwdlE}{Office-31}~\cite{DBLP:conf/eccv/SaenkoKFD10} and \href{http://ai.bu.edu/visda-2017/}{VisDA-2017}~\cite{peng2017visda}.
 From our experiments, we show that by jointly training contradistinguisher on the source domain and the target domain distributions, we can achieve above/on-par results over several recently proposed domain adaptation methods. 
 {We also observed an improvement in the classification performance on {VisDA-2017}~\cite{peng2017visda} over the vanilla CUDA with the data augmentation.}
} 
\item We further demonstrate our proposed method's simplicity and effectiveness by easily extending single-source domain adaptation to a more complex and general multi-source domain adaptation. We demonstrate the effectiveness of the multi-source domain adaptation extension by performing experiments on Office-31~\cite{DBLP:conf/eccv/SaenkoKFD10} dataset and \textcolor{black}{Digits datasets ( USPS (\href{https://web.stanford.edu/~hastie/ElemStatLearn//datasets/}{$us$})~\cite{lecun1989backpropagation}, MNIST (\href{http://yann.lecun.com/exdb/mnist/}{$mn$})~\cite{lecun1998gradient}, SVHN (\href{http://ufldl.stanford.edu/housenumbers/}{$sv$})~\cite{37648}, MNIST-M (\href{https://drive.google.com/file/d/0B9Z4d7lAwbnTSVR1dEFSRUFxOUU/view}{$mm$})~\cite{pmlr-v37-ganin15}, SYNNUMBERS (\href{https://drive.google.com/file/d/0B9Z4d7lAwbnTSVR1dEFSRUFxOUU/view}{$sn$})~\cite{pmlr-v37-ganin15}) in a multi-source setting.}
\item \color{black}\color{black}    Apart from these real-world benchmark datasets, we also validate the proposed method using the synthetically created toy-datasets (Fig.~\ref{fig:toy dataset demo small}). From our toy-dataset experiments, we provide two main insights.

\noindent
\begin{inparaenum}[(i)]
{\item \smallmethodname{} does indeed address the classification directly on the target domain in a fully unsupervised way without the domain alignment.}

\noindent
{\item Since the classification is done directly on the unlabeled target domain in a fully unsupervised manner, the target domain classification performance is not limited by the source domain classification performance, i.e., the irrespective of the domain is used as the labeled source domain and the unlabeled target domain, the performance is the respective domains are similar. In other words, swapping of the domains or the direction of the domain adaptation has little effect on the classification performance on each individual domain.
}\end{inparaenum}
\color{black}
\end{enumerate}

The rest of this paper is structured as follows. Section \ref{sec:literature review} discusses related works in domain adaptation. In Section \ref{sec:problem formulation}, we elaborate on the problem formulation, neural network architecture used by us, loss functions, model training, and inference algorithms of our proposed method. Section \ref{sec:experiments} deals with the discussion of the experimental setup, results and analysis on visual datasets. Finally, in Section \ref{sec:conclusions and future work}, we conclude by highlighting the key contributions of \smallmethodname{}.

\section{Related Work} \label{sec:literature review}
As mentioned earlier, almost all domain adaptation approaches rely on domain alignment techniques. Here we briefly outline three main techniques of domain alignment.

\begin{inparaenum}[(a)]
\item \emph{Discrepancy-based methods}:
Deep Adaptation Network (DAN)~\cite{DBLP:conf/icml/LongC0J15} proposes mean-embedding matching of multi-layer representations across domain by minimizing Maximum Mean Discrepancy (MMD)~\cite{Gretton:2009:FCK:2984093.2984169, gretton2012kernel, sejdinovic2013equivalence} in a reproducing kernel Hilbert space (RKHS).
Residual Transfer Network (RTN)~\cite{DBLP:conf/nips/LongZ0J16} introduces separate source and target domain classifiers differing by a small residual function along with fusing the features of multiple layers in a reproducing kernel Hilbert space (RKHS) to match the domain distributions.
Joint Adaptation Network (JAN)~\cite{DBLP:conf/icml/LongZ0J17} proposes to optimize Joint Maximum Mean Discrepancy (JMMD), which measures the Hilbert-Schmidt norm between kernel mean embedding of empirical joint distributions of source and target domain.
Associative Domain Adaptation (ADA)~\cite{DBLP:conf/iccv/HausserFMC17} learns statistically domain invariant embeddings by associating the embeddings of the final fully-connected layer before applying softmax as an alternative to MMD loss.
Maximum Classifier Discrepancy (MCD)~\cite{8578490} aligns source and target distributions by maximizing the discrepancy between two separate classifiers.
Self Ensembling (SE)~\cite{french2018selfensembling} uses mean teacher variant~\cite{DBLP:conf/nips/TarvainenV17} of temporal ensembling~\cite{DBLP:conf/iclr/LaineA17} with heavy reliance on data augmentation to minimize the discrepancy between student and teacher network predictions.
Variational Fair Autoencoder (VFAE)~\cite{DBLP:journals/corr/LouizosSLWZ15} uses Variational Autoencoder (VAE)~\cite{DBLP:journals/corr/KingmaW13} with MMD to obtain domain invariant features.
Central Moment Discrepancy (CMD)~\cite{2017arXiv170208811Z} proposes to match higher-order moments of source and target domain distributions. 
Rozantsev et al. ~\cite{rozantsev2018beyond} propose to explicitly model the domain shift using two-stream architecture, one for each domain along with MMD to align the source and target representations.
A more recent approach multi-domain Domain Adaptation layer (mDA-layer)~\cite{8792192, mancini2018boosting} proposes a novel idea of replacing standard Batch-Norm layers~\cite{ioffe2015batch} with specialized Domain Alignment layers~\cite{cariucci2017autodial, carlucci2017just} thereby reducing the domain shift by discovering and handling multiple latent domains.
Geodesic Flow Subspaces (GFS/SGF)~\cite{gopalan2011domain} performs domain adaptation by first generating two subspaces of the source and the target domains by performing PCA, followed by learning a finite number of the interpolated subspaces between source and target subspaces based on the geometric properties of the Grassmann manifold. In the presence of multi-source domains, this method is very effective as this identifies the optimal subspace for domain adaptation.
sFRAME (sparse Filters, Random fields and Maximum Entropy)~\cite{xie2015learning} models are defined as Markov random field model that model data distributions based on maximum entropy distribution to fit the observed data by identifying the patterns in the observed data. 
{
Transferrable Prototypical Networks (TPN)~\cite{DBLP:conf/cvpr/PanYLWNM19} propose to identify prototypes for each class in source and target domains that are close in the embedding space and minimize the distance between these prototypes for domain adaptation.}
{Contrastive Adaptation Network (CAN)~\cite{DBLP:conf/cvpr/Kang0YH19} uses MMD-loss for the feature encodings along with the heuristic clustering schema to selectively pick a subset of the high confidence image samples from the target domain. These samples are then utilized in training the classifier on the target domain instead of the entire target domain.}
\textcolor{black}{Moment Matching for Multi-Source Domain Adaptation (M3SDA)~\cite{DBLP:conf/iccv/PengBXHSW19} proposes to dynamically align multiple labeled source domains and the unlabeled target domain by matching the moments of the feature distributions.}

\item \emph{Reconstruction-based methods}:
Deep Reconstruction-Classification Networks (DRCN)~\cite{10.1007/978-3-319-46493-0_36} and Domain Separation Networks (DSN)~\cite{Bousmalis:2016:DSN:3157096.3157135} approaches learn shared encodings of source and target domains using reconstruction networks.

\item \emph{Adversarial adaptation methods}:
Reverse Gradient (RevGrad)~\cite{pmlr-v37-ganin15} or Domain Adversarial Neural Network (DANN)~\cite{ganin2016domain} uses domain discriminator to learn domain invariant representations of both the domains.
Coupled Generative Adversarial Network (CoGAN)~\cite{NIPS2016_6544} uses Generative Adversarial Network (GAN)~\cite{Goodfellow:2014:GAN:2969033.2969125} to obtain domain invariant features used for classification.
Adversarial Discriminative Domain Adaptation (ADDA)~\cite{8099799} uses GANs along with weight sharing to learn domain invariant features.
Generate to Adapt (GTA)~\cite{DBLP:conf/cvpr/Sankaranarayanan18a} learns to generate an equivalent image in the other domain for a given image, thereby learning common domain invariant embeddings. 
Cross-Domain Representation Disentangler (CDRD)~\cite{DBLP:conf/cvpr/LiuYFWCW18} learns cross-domain disentangled features for domain adaptation.
Symmetric Bi-Directional Adaptive GAN (SBADA-GAN)~\cite{Russo_2018_CVPR} aims to learn symmetric bidirectional mappings among the domains by trying to mimic a target image given a source image. 
Cycle-Consistent Adversarial Domain Adaptation (CyCADA)~\cite{pmlr-v80-hoffman18a} adapts representations at both the pixel-level and
feature-level over the domains.
Moving Semantic Transfer Network (MSTN)~\cite{xie2018learning} proposes a moving semantic transfer network that learns semantic representations for the unlabeled target samples by aligning labeled source centroids and pseudo-labeled target centroids.
Conditional Domain Adversarial Network (CDAN)~\cite{NIPS2018_7436} conditions the adversarial adaptation models on discriminative information conveyed in the classifier predictions.
Decision-boundary Iterative Refinement Training with a Teacher (DIRT-T)~\cite{shu2018a} and Augmented Cyclic Adversarial Learning (ACAL)~\cite{hosseini-asl2018augmented} learn by using a domain discriminator along with data augmentation for domain adaptation.
Deep Cocktail Network (DCTN)~\cite{xu2018deep} proposes a k-way domain discriminator
and category classifier for digit classification and real-world
object recognition in a multi-source domain adaptation setting.
{Batch Spectral Penalization (BSP)~\cite{chen2019transferability} investigates the transferability and the discriminability of the features learnt using the standard adversarial domain adaptation techniques. Also, BSP proposes an additional batch spectral loss as it is observed that the transferable features learnt using adversarial domain adaptation result in the loss of the discriminability of the classifier.
Transferable Normalization (TransNorm)~\cite{wang2019transferable} proposes a further improvement in transferability by replacing the normal batch-normalization layer with separate normalization layers for source and target domain input batches.
Adversarial Tight Match (ATM)~\cite{9080115} proposes to combine the adversarial training with discrepancy metric by introducing a novel discrepancy metric Maximum Density Divergence (MDD) to minimize inter-domain divergence and maximize the intra-class density.
Certainty based Attention for Domain Adaptation (CADA)~\cite{DBLP:conf/cvpr/KurmiKN19} propose to identify features that increase the certainty of the domain discriminator prediction to improve the classifier.
Progressive Feature Alignment Network (PFAN)~\cite{DBLP:conf/cvpr/ChenXHRD0XH19
} proposes to align the discriminative features across domains progressively and effectively, via exploiting the intra-class variation in the target domain. 
}
\end{inparaenum}

Apart from these approaches, a slightly different method that has been recently proposed is called \emph{Tri-Training}. Tri-Training algorithms use three classifiers trained on the labeled source domain and refine them for the unlabeled target domain. To be precise, in each round of tri-training, a target sample is pseudo-labeled if the other two classifiers agree on the labeling, under certain conditions such as confidence thresholding. Asymmetric Tri-Training (ATT)~\cite{pmlr-v70-saito17a} uses three classifiers to bootstrap high confidence target domain samples by confidence thresholding. This way of bootstrapping works only if the source classifier has very high accuracy. In the case of low source classifier accuracy, target samples are never obtained to bootstrap, resulting in a bad model. 
Multi-Task Tri-training (MT-Tri)~\cite{DBLP:conf/acl/PlankR18} explores the tri-training technique on the language domain adaptation tasks in a multi-task setting.

All the domain adaptation approaches mentioned earlier have a common unifying theme: they attempt to morph the target and source distributions so as to make them indistinguishable. However, aligning domains is a complex task than the classification task.
In this paper, we propose a completely different approach: instead of focusing on aligning the source and target distributions, we learn a single classifier referred to as \emph{Contradistinguisher}, jointly on both the domain distributions using contradistinguish loss for the unlabeled target domain data and supervised loss for the labeled source domain data. 

%
%
%
%

 


\section{Proposed Method: \smallmethodname{}} \label{sec:problem formulation}
A domain $\mathcal{D}_d$ is specified by its input feature space $\mathcal{X}_d$, the label space $\mathcal{Y}_d$ and the joint probability distribution $p(\fatx_d, \faty_d)$, where $\fatx_d{ \in }\mathcal{X}_d$ and $\faty_d{ \in }\mathcal{Y}_d$. Let $\left \lvert \mathcal{Y}_d \right \rvert{ = }K$ be the number of class labels such that $\faty_d{ \in }\{0, \ldots, K{ - }1\}$ for any instance $\fatx_d$. Domain adaptation, in particular, consists of two domains $\mathcal{D}_{s}$ and $\mathcal{D}_{t}$ that are referred as the source and target domains respectively. 
{We define $(\fatx_s,\faty_s)$ as the random variables that denote the source domain input features and the corresponding source domain label. Similarly, we define $(\fatx_t,\faty_t)$ as the random variables that denote the target domain input features and the corresponding target domain label. }
A common assumption in domain adaptation is that the input feature space as well as the label space remains unchanged across the source and the target domain, i.e., $\mathcal{X}_s{ = }\mathcal{X}_t{ = }\mathcal{X}_d$ and $\mathcal{Y}_s{ = }\mathcal{Y}_t{ = }\mathcal{Y}_d$. Hence, the only difference between the source and target domain is input-label space distributions, i.e., $p(\fatx_s,\faty_s){ \neq }p(\fatx_t,\faty_t)$. This is referred to as \emph{domain shift} in the domain adaptation literature.

{In particular, in an unsupervised domain adaptation, the training data consists of the labeled source domain instances ${\{(\fatx^{i}_s, \faty^{i}_s)\}}^{n_s}_{i=1}$ corresponding to the random variables $(\fatx_s,\faty_s)$ and the unlabeled target domain instances ${\{\fatx^{j}_t\}}^{n_t}_{j=1}$ corresponding to the random variables $(\fatx_t)$. Observe that in the unsupervised domain adaptation setting, the target domain labels ${\{\faty^{j}_t\}}^{n_t}_{j=1}$ corresponding to the random variable $(\faty_t)$ are unobserved/missing.} Given a labeled data in the source domain, it is straightforward to learn a classifier by maximizing the conditional probability $p(\faty_s|\fatx_s)$ over the labeled samples. However, the task at hand is to learn a classifier on the unlabeled target domain by transferring the knowledge from the labeled source domain to the unlabeled target domain.

\subsection{Overview} \label{sec:overview}
\begin{figure}[t!]
\centering
\includegraphics[width=\linewidth,height=\textheight,keepaspectratio=true]{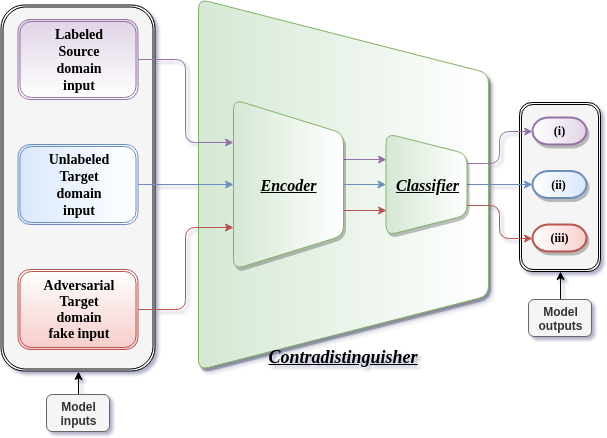}
\caption{Architecture of the proposed method \smallmethodname{} with \emph{Contradistinguisher} (\emph{Encoder} and \emph{Classifier}). Three optimization objectives with their respective inputs involved in training of \smallmethodname{}: (i) Source supervised~\eqref{eq:ce}, (ii) Target unsupervised~\eqref{eq:unsup} and Adversarial regularization~\eqref{eq:adv reg loss}.}
\label{fig:training method}
\end{figure}
The outline of the proposed method CUDA that involves \emph{contradistinguisher} and the respective losses involved in training are depicted in Fig.~\ref{fig:training method}. 
The objective of contradistinguisher is to find a clustering scheme using the most contrastive features on unlabeled target in such a way that it also satisfies the {target domain prior over the labels, i.e., \emph{target domain prior enforcing}}. We achieve this by jointly training on labeled source samples in a supervised manner and unlabeled target samples in an unsupervised end-to-end manner by using a contradistinguish loss same as~\cite{Pandey2017UnsupervisedFL}. 

This fine-tunes the classifier learnt from the source domain also to the target domain, as demonstrated in Fig.~\ref{fig:toy dataset demo small}. The crux of our approach is 
the contradistinguish loss~\eqref{eq:unsup} which is discussed in detail in Section~\ref{sec:unsupervised target classification}. Hence, the apt name \emph{contradistinguisher} for our neural network architecture.

Note that the objective of contradistinguisher is not the same as a classifier, i.e., distinguishing is not the same as classifying.
Suppose there are two contrastive entities $e_1{ \in }C_1$ and $e_2{ \in }C_2$, where $C_1, C_2$ are two classes. The aim of a classifier is to classify $e_1{ \in }C_1$ and $e_2{ \in }C_2$, where to train a classifier one requires labeled data. On the contrary, the job of contradistinguisher is to just identify $e_1{ \neq }e_2$, i.e., contradistinguisher can classify $e_1{ \in }C_1$ (or $C_2$) and $e_2{ \in }C_2$ (or $C_1$) indifferently. To train contradistinguisher, we do not need any class information but only need unlabeled entities $e_1$ and $e_2$. Using unlabeled target data, contradistinguisher is able to find a clustering scheme by distinguishing the unlabeled target domain samples in an unsupervised way. However, since the final task is classification, one would require a selective incorporation of the pre-existing informative knowledge required for the task of classification. This knowledge of assigning the label to the clusters is obtained by jointly training, thus classifying $e_1{ \in }C_1$ and $e_2{ \in }C_2$.

\subsection{Supervised Source Classification} \label{sec:supervised source classification}
For the labeled source domain instances ${\{(\fatx^{i}_s, \faty^{i}_s)\}}^{n_s}_{i=1}$ {corresponding to the random variables $(\fatx_s,\faty_s)$ of the labeled source domain}, we define the conditional-likelihood of observing $\faty_s$ given $\fatx_s$ as, $p_{\theta}(\faty_{s}|\textbf{x}_{s})$,
where $\theta$ denotes the parameters of contradistinguisher.

We estimate $\theta$ by maximizing the conditional log-likelihood of observing the labels given the labeled source domain samples. Therefore, the source domain supervised objective to maximize is given as
\setlength{\arraycolsep}{0.0em}
\begin{eqnarray}
\mathcal{L}_{s}(\theta)&{}={}&\sum_{i=1}^{n_s}\log(p_{\theta}(\faty^{i}_{s}|\textbf{x}^{i}_{s}))\enspace.
\label{eq:sup}
\end{eqnarray}
\setlength{\arraycolsep}{5pt}
Alternatively, one can minimize the cross-entropy loss, as used in practical implementation, instead of maximizing~\eqref{eq:sup}, i.e.,
\setlength{\arraycolsep}{0.0em}
\begin{eqnarray}
\mathcal{L}_{ce}(\theta)&{}={}&-\sum_{i=1}^{n_s}\sum_{k=0}^{K-1}{\mathbbm{1}[\faty^{i}_s{ = }k]}\log(\hat{\faty}^{ik}_{s})\enspace,
\label{eq:ce}
\end{eqnarray}
where 
$\hat{\faty}^{ik}_{s}$ is the softmax output of contradistinguisher that represents the probability of class $k$ for the given sample $\fatx^{i}_s$.

\subsection{Unsupervised Target Classification} \label{sec:unsupervised target classification}
For the unlabeled target domain instances ${\{\fatx^{j}_t\}}^{n_t}_{j=1}$ {corresponding to the random variable $\fatx_t$ of the unlabeled target domain, the corresponding labels ${\{\faty^{j}_t\}}^{n_t}_{j=1}$ corresponding to the random variable $\fatx_t$ are unknown/missing}. Hence, a naive way of predicting the target labels is to directly use the classifier trained only with a supervised loss given in~\eqref{eq:ce}. While this approach may perform reasonably well in certain cases, it fails to deliver state-of-the-art performance. This may be attributed to the following reason:
the support for the distribution $p_{\theta}$ is defined only over the source domain instances $\fatx_s$ and not the target domain instances $\fatx_t$.
Hence, we model a non-trivial joint distribution $\hat{q}_\theta(\textbf{x}_t,\faty_t)$ parameterized by the same $\theta$ over target domain with only the target domain instances as the support as,
\setlength{\arraycolsep}{0.0em}
\begin{eqnarray}
\hat{q}_{\theta}(\textbf{x}_t,\faty_t)&{}={}&\frac{p_{\theta}(\faty_t|\textbf{x}_t)}{\sum_{\ell=1}^{n_t}p_{\theta}(\faty_t|\textbf{x}_t^{\ell})}\enspace.
\label{eq:mod unsup1}
\end{eqnarray}
\setlength{\arraycolsep}{5pt}
{However~\eqref{eq:mod unsup1} is not a joint distribution yet because $\sum_{\ell=1}^{n_t}\hat{q}_{\theta}(\textbf{x}^{\ell}_{t}, \faty_{t}){ \neq }p(\faty_t)$, i.e., marginalizing over all ${\{\fatx^{j}_t\}}^{n_t}_{j=1}$ does not yield the \emph{target prior distribution}, i.e., $p(\faty_t)$. }
We modify~\eqref{eq:mod unsup1} so as to include the marginalization condition. Hence, we refer to this as \emph{target domain prior enforcing}.
\setlength{\arraycolsep}{0.0em}
\begin{eqnarray}
q_{\theta}(\textbf{x}_t,\faty_t)&{}={}&\frac{p_{\theta}(\faty_t|\textbf{x}_t) {p(\faty_t)}}{\sum_{\ell=1}^{n_t}p_{\theta}(\faty_t|\textbf{x}_t^{\ell})}\enspace,
\label{eq:mod unsup}
\end{eqnarray}
\setlength{\arraycolsep}{5pt}
{where $p(\faty_t)$ denotes the target domain prior probability over the labels.}

Note that $q_\theta(\fatx_t, \faty_t)$ defines a non-trivial approximate of joint distribution over the target domain as a function of $p_\theta$ learnt over source domain. The resultant unsupervised maximization objective for the target domain is given by maximizing the log-probability of the joint distribution $q_\theta(\fatx_t, \faty_t)$ which is
\setlength{\arraycolsep}{0.0em}
\begin{eqnarray}
\mathcal{L}_{t}(\theta, {\{\faty^{j}_t\}}^{n_t}_{j=1})&{}={}&\sum_{j=1}^{n_t}\log(q_{\theta}(\textbf{x}^{j}_{t}, \faty^{j}_{t}))\enspace,
\label{eq:unsup}
\end{eqnarray}
\setlength{\arraycolsep}{5pt}
Next, we discuss how the objective given in~\eqref{eq:unsup} is solved, and the reason why~\eqref{eq:unsup} is referred to as contradistinguish loss.
Since the target labels ${\{\faty^{j}_t\}}^{n_t}_{j=1}$ are unknown, one needs to maximize~\eqref{eq:unsup} over the parameters $\theta$ as well as the unknown target labels $\faty_t$. As there are two unknown variables for maximization, we follow a two-step approach to maximize~\eqref{eq:unsup} as analogous to Expectation-Maximization (EM) algorithm~\cite{dempster1977maximum}. The two optimization steps are as follows.

\begin{inparaenum}[(i)] 
\item \emph{Pseudo-label selection}: We maximize \eqref{eq:unsup} only with respect to the label $\faty_t$ for every $\fatx_t$ by fixing $\theta$ as
\setlength{\arraycolsep}{0.0em}
\begin{eqnarray}
\hat{\faty}^{j}_t&{}={}&\argmax_{\faty^{j} \in \mathcal{Y}_t} \frac{p_{\theta}(\faty^{j}|\textbf{x}^{j}_t) {p(\faty_t)}}{\sum_{\ell=1}^{n_t}p_{\theta}(\faty^{\ell}| \textbf{x}_t^{\ell})}\enspace,
\label{eq:pseudo label selection}
\end{eqnarray}
\setlength{\arraycolsep}{5pt}
\noindent
Pseudo-labeling approach under semi-supervised representation learning setting has been well studied in~\cite{pseudo-label} and shown equivalent to \emph{entropy regularization}~\cite{grandvalet2005semi}. As previously mentioned, the pseudo-label selection is analogous to E-step in the EM algorithm. Moreover, we derive the motivation from~\cite{Pandey2017UnsupervisedFL} that also uses pseudo-labeling in the context of semi-supervised representation learning. However, the proposed method addresses a more complex problem of domain adaptation in the presence of a domain shift. The pseudo-labeling essentially tries to cluster by assigning labels using source domain features of the classifier trained on the source domain. This is effectively is similar to the E-step in EM algorithm in spirit.

\item \emph{Maximization}: By fixing the pseudo-labels ${\{\hat{\faty}^{j}_t\}}^{n_t}_{j=1}$ from~\eqref{eq:pseudo label selection}, we train contradistinguisher to maximize \eqref{eq:unsup} with respect to the parameter $\theta$.
\setlength{\arraycolsep}{0.0em}
\begin{eqnarray}
\mathcal{L}_t(\theta)&{}={}&\sum_{j=1}^{n_t}\log(p_{\theta}(\hat{\faty}^{j}_t|\textbf{x}^{j}_t)) + \sum_{j=1}^{n_t}\log({p(\faty_t)})\nonumber\\
&&{-}\:\sum_{j=1}^{n_t}\log(\sum_{\ell=1}^{n_t}p_{\theta}(\hat{\faty}^{j}_t| \textbf{x}_t^{\ell}))\enspace.
\label{eq:unsup full}
\end{eqnarray}
\setlength{\arraycolsep}{5pt}
\noindent
Since the pseudo-labels from~\eqref{eq:pseudo label selection} are used for the maximization, this constrains the model to learn the features to further improve the current pseudo-labeling for the next iteration. This step is similar to the M-step in the EM algorithm in spirit.

The first term, i.e., log-probability for a label $\hat{\faty}^{j}_t$ given $\fatx^{j}_t$ forces contradistinguisher to choose features to classify $\fatx^{j}_t$ to $\hat{\faty}^{j}_t$. The second term is a constant, hence it has no effect on the optimization with respect to $\theta$. {The third term is the negative of log-probability for the pseudo-label $\hat{\faty}^{j}_t$ given all the samples $\fatx^{\ell}_t$ in the entire domain. Maximization of this term forces contradistinguisher to choose features to not classify all the other $\fatx^{\ell{ \neq }j}_t$ to selected pseudo-label $\hat{\faty}^{j}_t$ except the given sample $\fatx^{j}_t$.} This forces contradistinguisher to extract the most unique features of a given sample $\fatx^{j}_t$ against all the other samples $\fatx^{\ell{ \neq }j}_t$, i.e., most unique contrastive feature of the selected sample $\fatx^{j}_t$ over all the other samples $\fatx^{\ell{ \neq }j}_t$ to distinguish a given sample from all others.

The first and third term together in~\eqref{eq:unsup full} enforce that contradistinguisher learns the most contradistinguishing features among the samples $\fatx_t{ \in }\mathcal{X}_t$, thus performing unlabeled target domain classification in a fully unsupervised way. We refer to the unsupervised target domain objective \eqref{eq:unsup} as contradistinguish loss because of this contradistinguishing feature learning.
\end{inparaenum}

Ideally, one would like to compute the third term in~\eqref{eq:unsup full} using the complete target training data for each input sample. Since it is expensive to compute the third term over the entire $\fatx_t$ for each individual sample during training, one evaluates the third term in~\eqref{eq:unsup full} over a mini-batch. In our experiments, we have observed that the mini-batch strategy does not cause any problem during training as far as it includes at least one sample from each class, which is a fair assumption for a reasonably large mini-batch size of $128$. 
For numerical stability, we use $\log\sum\exp$ trick to optimize third term in~\eqref{eq:unsup full}.

\subsection{Adversarial Regularization} \label{sec:adversarial regularization}
To prevent contradistinguisher from over-fitting to the chosen pseudo labels during the training, we use adversarial regularization. In particular, we train contradistinguisher to be confused about the set of fake negative samples ${\{\hat{\fatx}^{j}_t\}}^{n_f}_{j=1}$ by maximizing the conditional log-probability over the given fake sample such that the sample belongs to all $K(\left \lvert \mathcal{Y}_d \right \rvert)$ classes simultaneously. The adversarial regularization objective is to multi-label the fake sample (e.g., a noisy image that looks like a cat and a dog) equally to all $K$ classes as labeling to any unique class introduces more noise in pseudo labels. This strategy is similar to entropy regularization~\cite{grandvalet2005semi} in the sense that instead of minimizing the entropy for the real target samples, we maximize the conditional log-probability over the fake negative samples. Therefore, we add the following maximization objective to the total contradistinguisher objective as a regularizer. 
\setlength{\arraycolsep}{0.0em}
\begin{eqnarray}
\mathcal{L}_{adv}(\theta)&{}={}&\sum_{j=1}^{n_f}\log(p_{\theta}(\hat{\faty}_t^{j}|\hat{\fatx}^{j}_{t}))\enspace,
\label{eq:adv reg}
\end{eqnarray}
for all $\hat{\faty}^{j}_t{ \in }\mathcal{Y}_t$.
As maximization of \eqref{eq:adv reg} is analogous to minimizing the binary cross-entropy loss \eqref{eq:adv reg loss} of a multi-class multi-label classification task, in our practical implementation, we minimize \eqref{eq:adv reg loss} for assigning labels to all the classes for every sample.
\setlength{\arraycolsep}{0.0em}
\begin{eqnarray}
\mathcal{L}_{bce}(\theta)&{}={}&-\sum_{j=1}^{n_f}\sum_{k=0}^{K-1}\log(\hat{\faty}^{jk}_{t})\enspace,
\label{eq:adv reg loss}
\end{eqnarray}
where 
$\hat{\faty}^{jk}_{t}$ is the softmax output of contradistinguisher which represents the probability of class $k$ for the given sample $\hat{\fatx}^{j}_t$.

The fake negative samples $\hat{\fatx}_t$ can be directly sampled from, say, a Gaussian distribution in the input feature space $\mathcal{X}_t$ with the mean and standard deviation of the samples $\fatx_t{ \in }\mathcal{X}_t$. For the language domain, fake samples are generated randomly, as mentioned above, because the input feature is the form of embeddings extracted from denoising auto-encoder with bag-of-words as the input. 
In case of visual datasets, as the feature space is high dimensional, the fake images $\hat{\fatx}_t$ are generated using a generator network $G_{\phi}$ with parameter $\phi$ that takes Gaussian noise vector $\eta_t$ as input to produce a fake sample $\hat{\fatx}_t$, i.e., $\hat{\fatx}_t = G_{\phi}(\eta_t)$. Generator $G_{\phi}$ is trained by minimizing {kernel-MMD} loss~\cite{DBLP:conf/nips/LiCCYP17}, i.e., a modified version of MMD loss between the encoder output $\rho_{enc}(\hat{\fatx}_t)$ and $\rho_{enc}(\fatx_t)$ of $n_f$ fake images $\hat{\fatx}_t$ and $n_t$ real target domain images $\fatx_t$ respectively. 
\setlength{\arraycolsep}{0.0em}
\begin{eqnarray}
\mathcal{L}_{gen}(\phi)&{}={}&\frac{1}{n_f^{2}}\sum_{i=1}^{n_f}\sum_{j=1}^{n_f} k(\rho_{enc}(\hat{\fatx}_t^{i}),\rho_{enc}(\hat{\fatx}_t^{j}))\nonumber\\
&&{+}\:\frac{1}{n_t^{2}}\sum_{i=1}^{n_t}\sum_{j=1}^{n_t} k(\rho_{enc}(\fatx_t^{i}),\rho_{enc}(\fatx_t^{j}))\nonumber\\
&&{-}\:\frac{2}{n_t n_f}\sum_{i=1}^{n_f}\sum_{j=1}^{n_t} k(\rho_{enc}(\hat{\fatx}_t^{i}),\rho_{enc}(\fatx_t^{j})),
\label{eq:kMMD loss gen}
\end{eqnarray}
\setlength{\arraycolsep}{5pt}
\noindent where $k(x,x') = e^{-\gamma\norm{x-x'}^{2}}$ is the Gaussian kernel.
 
Note that the generator's objective is not to generate realistic images but to generate fake noisy images with mixed image attributes from the target domain. This reduces the effort of training powerful generators, which is the focus in adversarial based domain adaptation approaches~\cite{DBLP:conf/cvpr/Sankaranarayanan18a, DBLP:conf/cvpr/LiuYFWCW18, Russo_2018_CVPR, pmlr-v80-hoffman18a, xie2018learning} used for domain alignment.
Algorithm~\ref{alg: method training} and~\ref{alg: method inference} list steps involved in \smallmethodname{} training and inference, respectively. 
\begin{algorithm}[t!]
\DontPrintSemicolon
  
  \KwInput{$b{ = }batch\_size$, $epochs{ = }max\_epoch$, $n_{batch}{ = } number\ of\ batches$}
  \KwOutput{$\theta$ \color{black}{{/* parameter of contradistinguisher */}}}
  \KwData{${\{(\fatx^{i}_s, \faty^{i}_s)\}}^{n_s}_{i=1}$, ${\{\fatx^{j}_t\}}^{n_t}_{j=1}$}
    \If{target domain prior $p(\faty_t)$ is known}
    {
        use $p(\faty_t)$ for the contradistinguish loss~\eqref{eq:unsup}  \newline{/* target domain prior enforcing */}
    }
    \Else
    {
    	compute $p(\faty_t)$ assuming $p(\faty_t) = p(\faty_s)$  \newline{/* fair assumption as most datasets are well balanced */}
	}
    \For{$epoch = 1$ to $epochs$}
    {
        \For{$batch = 1$ to $n_{batch}$}
        {
            sample a mini-batch ${\{(\fatx^{i}_s, \faty^{i}_s)\}}^{b}_{i=1}$, ${\{\fatx^{j}_t\}}^{b}_{j=1}$
            
            compute $\mathcal{L}_{s}(\theta)$~\eqref{eq:sup}\label{alg_stt: source supervised loss} using ${\{(\fatx^{i}_s, \faty^{i}_s)\}}^{b}_{i=1}$ \newline{/* source supervised loss */}
            
            compute ${\{\hat{\faty}^{j}_t\}}^{b}_{j=1}$~\eqref{eq:pseudo label selection} using ${\{\fatx^{j}_t\}}^{b}_{j=1}$  \newline{/* \emph{pseudo label selection} step */}\label{alg_stt: pseudo label selection}
            
            compute $\mathcal{L}_{t}(\theta)$~\eqref{eq:unsup full} fixing ${\{\hat{\faty}^{j}_t\}}^{b}_{j=1}$ \newline{/* \emph{maximization} step */}\label{alg_stt: target unsupervised loss}\label{alg_stt: maximization}
            \newline{/* steps \ref{alg_stt: pseudo label selection} and \ref{alg_stt: maximization} together optimize unsupervised contradistinguish loss~\eqref{eq:unsup} */}
            
            \If{adversarial regularization is enabled}
            {
                \If{Generator $G_\phi$ is used}
                {
                get fake samples ${\{\hat{\fatx}^{j}_t\}}^{b}_{j=1}$ from Gaussian noise vectors ${\{\eta^{j}_t\}}^{b}_{j=1}$ using $G_\phi$, compute $\mathcal{L}_{gen}(\phi)$\eqref{eq:kMMD loss gen}\label{alg_stt: generator loss}  \newline{/* generator training */}
                }
                \Else
                {
                get fake samples ${\{\hat{\fatx}^{j}_t\}}^{b}_{j=1}$ by random sampling in the input feature space $\mathcal{X}_t$
                }
                compute $\mathcal{L}_{adv}(\theta)$~\eqref{eq:adv reg loss} using ${\{\hat{\fatx}^{j}_t\}}^{b}_{j=1}$ \label{alg_stt: adv loss} \newline{/* fake samples are assigned to all classes equally */}
            }
            combine losses in steps~\ref{alg_stt: source supervised loss},\ref{alg_stt: target unsupervised loss},\ref{alg_stt: generator loss} and~\ref{alg_stt: adv loss} to compute gradients using back-propagation \label{alg_stt: total loss}  
            
            update $\theta$ using gradient descent \newline{/* and $\phi$ if $G_\phi$ is used */}
        }
    }
\caption{CUDA Training}
\label{alg: method training}
\end{algorithm}
\begin{algorithm}[t!]
\DontPrintSemicolon
  
  \KwInput{${\{\fatx^{i}_{test}\}}^{n_{test}}_{i=1}$  {/* input test samples */}}
  \KwOutput{${\{\hat{\faty}^{i}_{test}\}}^{n_{test}}_{i=1}$ {/* predicted labels */}}

    \For{$i = 1$ to $n_{test}$}
    {
        
        predict label as $\hat{\faty}^{i}_{test}{}={}\argmax_{\faty \in \mathcal{Y}_t} p_{\theta}(\faty|\textbf{x}^{i}_{test})$\label{alg_stt: predict labels}
    }
\caption{CUDA Inference}
\label{alg: method inference}
\end{algorithm}

\subsection{Extension to Multi-Source Domain Adaptation} \label{sec: multisource domain adaptation}
{Here, we argue that our proposed method CUDA has an implicit advantage in dealing with multi-source domain adaption problems over the techniques based on domain alignment. In a multi-source adaption setting, domain alignment methods need to consider the domain-shift between a source and a target domain and consider the domain-shift between the multiple source domains. Therefore, domain alignment methods are required to solve the even more complex intermediate problem of aligning multiple source and target domain distributions in addition to the complex intermediate problem of source and target domain alignment to deal with the multi-source domain adaptation problems.}  
{
However, as the proposed method does not depend on the domain alignment, the extension to multi-source in the proposed method is very simple.
As our main focus is to perform unsupervised learning directly on the target domain, the model obtained using is better generalized to the target domain and reduces overfitting on the source, which usually results in a negative transfer.
We believe that this is one of the main advantages of addressing the domain adaptation by performing the primary task of target domain classification rather than the intermediate task of domain alignment.
}

We propose a simple extension our proposed method to perform multi-source domain adaptation in the following manner. Let us suppose we are given with $R$ source domains ${\{s_1,\ldots,s_R\}}$, consisting of labeled training data $({\{(\fatx^{i}_{s_1}, \faty^{i}_{s_1})\}}^{n_{s_1}}_{i=1}, \ldots, {\{(\fatx^{i}_{s_R}, \faty^{i}_{s_R})\}}^{n_{s_R}}_{i=1})$ and unlabeled target domain instances ${\{\fatx^{j}_t\}}^{n_t}_{j=1}$.
We compute the source supervised loss for the $r^{th}$ source domain using~\eqref{eq:ce}, i.e., $\mathcal{L}_{s_r}(\theta)$~\eqref{eq:sup} with ${\{(\fatx^{i}_{s_r}, \faty^{i}_{s_r})\}}^{n_{s_r}}_{i=1}$ training data. We further compute the total multi-source supervised loss as \setlength{\arraycolsep}{0.0em}
\begin{eqnarray}
\mathcal{L}_{s_{total}}(\theta) = \sum_{r=1}^{R}\mathcal{L}_{s_r}(\theta).
\label{eq:sup multisource}
\end{eqnarray}
\setlength{\arraycolsep}{5pt}
We replace $\mathcal{L}_{s}(\theta)$~\eqref{eq:sup} in the total optimization objective with $\mathcal{L}_{s_{total}}(\theta)$~\eqref{eq:sup multisource} in step~\ref{alg_stt: total loss} of Algorithm~\ref{alg: method training}. It should be noted that the unsupervised loss for the target domain is still unmodified irrespective of the number of source domains. We experimentally demonstrate the efficacy of the proposed multi-source domain adaptation extension on Office-31~\cite{DBLP:conf/eccv/SaenkoKFD10} and \color{black}\color{black}Digits datasets~\cite{lecun1989backpropagation, lecun1998gradient, 37648, pmlr-v37-ganin15}.\color{black}

\section{Experiments} \label{sec:experiments}
{For our domain adaptation experiments, we consider both synthetic and real-world datasets. Under synthetic datasets, we experiment using {$2$-dimensional} blobs with different source and target domain probability distributions to demonstrate the effectiveness of the proposed method under different domain shifts. Under real-world datasets, we consider only the complex, high-resolution Office-31~\cite{DBLP:conf/eccv/SaenkoKFD10} and {VisDA-2017~\cite{peng2017visda}} object classification datasets for our experiment as the low-resolution datasets are already addressed in our conference paper CUDA: Contradistinguisher for Unsupervised Domain Adaptation (CUDA)~\cite{Balgi2019CUDA}.
We have published our python code for all the experiments at \url{https://github.com/sobalgi/cuda}.}

\color{black}\textcolor{black}
{
\begin{table}[t!]
\caption{Details of visual domain adaptation datasets.}
    \begin{center}
    \label{tbl:visual datasets}
    \setlength\tabcolsep{4pt}
    \begin{tabular}{llrcc}
        \toprule
    Dataset & Domain & \# Train & \# Test & \# Classes \\
    \midrule
     \multirow{3}{*}{Office-31} & AMAZON (\href{https://people.eecs.berkeley.edu/~jhoffman/domainadapt/}{$\mathcal{A}$})~\cite{DBLP:conf/eccv/SaenkoKFD10} & 2,817 & - & \multirow{3}{*}{31} \\ 
     & DSLR (\href{https://people.eecs.berkeley.edu/~jhoffman/domainadapt/}{$\mathcal{D}$})~\cite{DBLP:conf/eccv/SaenkoKFD10} & 498 & - &\\ 
     & WEBCAM (\href{https://people.eecs.berkeley.edu/~jhoffman/domainadapt/}{$\mathcal{W}$})~\cite{DBLP:conf/eccv/SaenkoKFD10} & 795 & - &\\ 
    \midrule
    \multirow{2}{*}{VisDA-2017} & {SYNTHETIC (\href{http://ai.bu.edu/visda-2017/}{$\mathcal{V}_{syn}$})~\cite{peng2017visda}} & {152,397} & {-} & \multirow{2}{*}{12}  \\ 
    & {REAL (\href{http://ai.bu.edu/visda-2017/}{$\mathcal{V}_{real}$})~\cite{peng2017visda}} & {55,388} & {72,372} &  \\ 
    \midrule
    \multirow{5}{*}{\color{black}\color{black}Digits\color{black}
} & USPS (\href{https://web.stanford.edu/~hastie/ElemStatLearn//datasets/}{$us$})~\cite{lecun1989backpropagation} & 7,291 & 2,007 & \multirow{5}{*}{10}\\ 
    & MNIST (\href{http://yann.lecun.com/exdb/mnist/}{$mn$})~\cite{lecun1998gradient} & 60,000 & 10,000 & \\ 
    & MNIST-M (\href{http://yann.lecun.com/exdb/mnist/}{$mm$})~\cite{pmlr-v37-ganin15} & 25,000 & 9,000 & \\ 
    & SVHN (\href{http://ufldl.stanford.edu/housenumbers/}{$sv$})~\cite{37648} & 73,257 & 26,032 &\\ 
    & SYNNUMBERS (\href{https://drive.google.com/file/d/0B9Z4d7lAwbnTSVR1dEFSRUFxOUU/view}{$sn$})~\cite{pmlr-v37-ganin15} & 479,400 & 9,553 &\\ 
    \bottomrule
    \end{tabular}
    \end{center}
\end{table}}
\color{black}
{Table~\ref{tbl:visual datasets} provides details on the visual datasets used in our experiments. 
We also experiment and report the results of our ablation study carried out with different combinations of the three optimization objectives with their respective domains as the inputs involved in \smallmethodname{} training: }

\noindent
\begin{inparaenum}[(i)]
    \noindent
    \item source supervised loss: $ss$ described in~\eqref{eq:ce},
    
    \noindent
    \item source/target unsupervised loss: $su/tu$ described in~\eqref{eq:unsup},
    
    \noindent
    \item source/target adversarial regularization loss: $sa/ta$ described in~\eqref{eq:adv reg loss}.
\end{inparaenum}

{$ss$ indicates the minimum target domain test accuracy that can be attained with a chosen contradistinguisher neural network by training only using the labeled source domain.
Any improvement over $ss$ using \smallmethodname{} (i.e., combination of $su/tu/sa/ta$) indicates effectiveness of \smallmethodname{} as the chosen contradistinguisher neural network is fixed. }

\subsection{Experiments on Synthetic Toy-dataset} \label{sec: toy exp setup}
\begin{figure*}[ht!]
\begin{center}
\subfloat[]{
\includegraphics[width=0.215\linewidth,height=0.215\textheight,keepaspectratio=true]{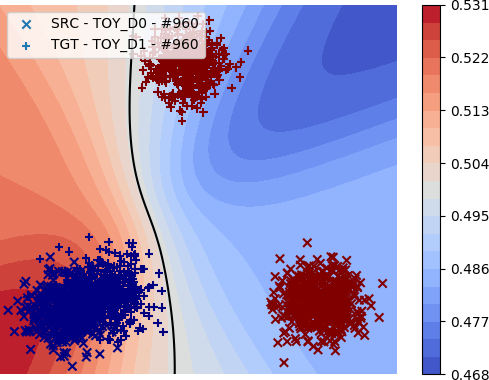}\label{sfig:cp_src_toy_d0_seed_1811_ss_tu_ta_contour_epoch}}
\hfill
\subfloat[]{
\includegraphics[width=0.215\linewidth,height=0.215\textheight,keepaspectratio=true]{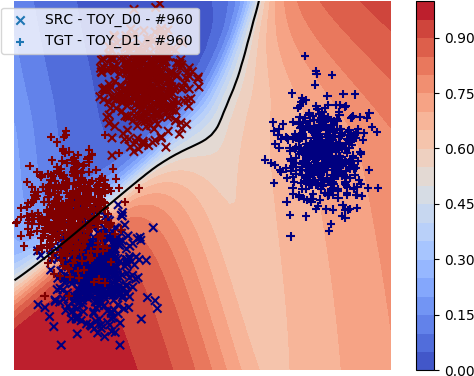}\label{sfig:src_toy_d0_seed_4019_ss_tu_ta_contour_epoch041}}
\hfill
\subfloat[]{
\includegraphics[width=0.215\linewidth,height=0.215\textheight,keepaspectratio=true]{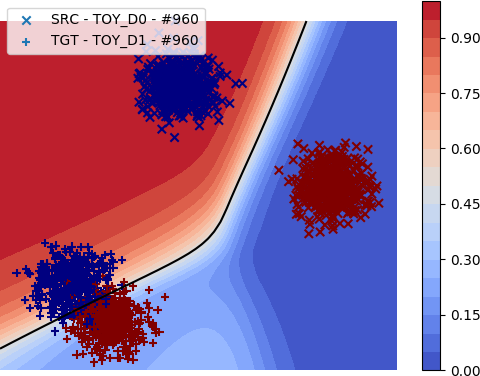}\label{sfig:src_toy_d0_seed_0925_ss_tu_ta_contour_epoch042}}
\hfill
\subfloat[]{
\includegraphics[width=0.215\linewidth,height=0.215\textheight,keepaspectratio=true]{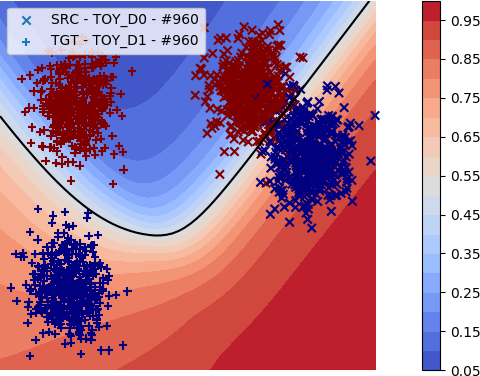}\label{sfig:src_toy_d0_seed_0661_ss_tu_ta_contour_epoch016}}%
\\
\subfloat[]{
\includegraphics[width=0.215\linewidth,height=0.215\textheight,keepaspectratio=true]{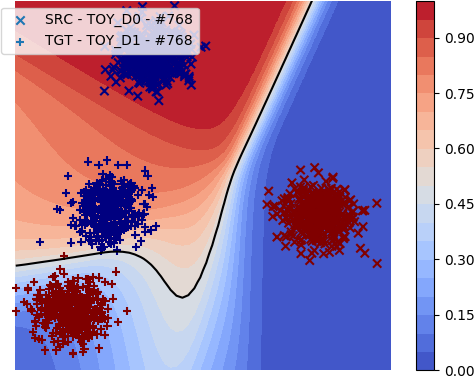}\label{sfig:src_toy_d0_seed_2108_ss_tu_ta_contour_epoch006}}
\hfill
\subfloat[]{
\includegraphics[width=0.215\linewidth,height=0.215\textheight,keepaspectratio=true]{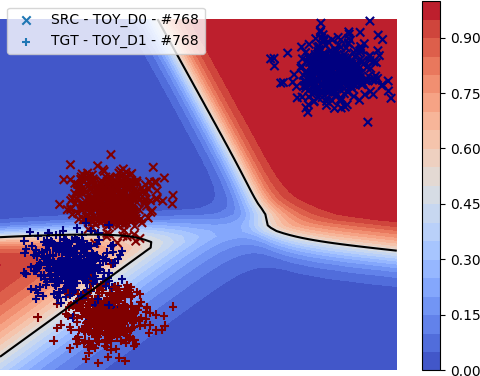}\label{sfig:src_toy_d0_seed_3182_ss_tu_ta_contour_epoch077}}
\hfill
\subfloat[]{
\includegraphics[width=0.215\linewidth,height=0.215\textheight,keepaspectratio=true]{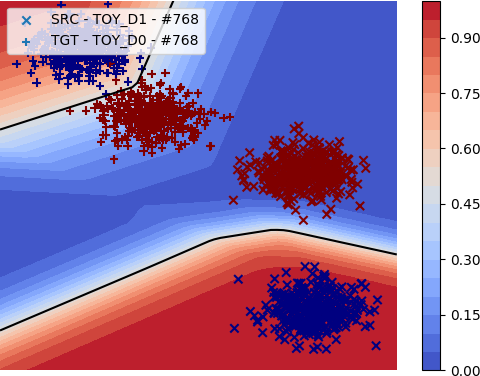}\label{sfig:src_toy_d1_seed_0152_ss_tu_ta_contour_epoch012}}
\hfill
\subfloat[]{
\includegraphics[width=0.215\linewidth,height=0.215\textheight,keepaspectratio=true]{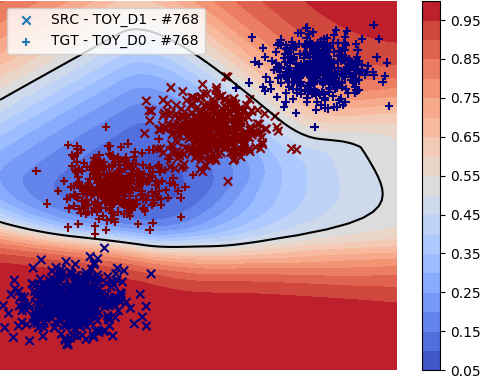}\label{sfig:src_toy_d1_seed_1887_ss_tu_ta_contour_epoch021}}
\\
\caption{
Contour plots show the probability contours along with clear decision boundaries on different toy-dataset settings trained using \smallmethodname{}.
(source domain: $\times$, target domain: $+$, class 0: blue, class 1: red.) 
(Best viewed in color.)
}
\label{fig:toy dataset plots}
\end{center}
\end{figure*}
We validate our proposed method by performing experiments on synthetically created simple datasets that model different source and target domain distributions in a {$2$-dimensional} input feature space using different blobs of source-target domain orientations and offsets (i.e., domain shift).
~We create blobs for source and target domains with 4000 samples using standard $scikit{ - }learn$~\cite{pedregosa2011scikit} as indicated in Fig.~\ref{fig:toy dataset demo small} and~\ref{fig:toy dataset plots}. We further evenly split these 4000 data-points into equal train and test sets. Each of the splits consists of the same number of samples corresponding to both the class labels. 

The main motivation of the experiments on toy-dataset is to understand and visualize the behavior of the proposed method under some typical domain distribution scenarios and analyze the performance of ~\smallmethodname{}.
$Blobs$ toy-dataset plots in Fig.~\ref{fig:toy dataset demo small} shows clear comparisons of the classifier decision boundaries learnt using \smallmethodname{} over domain alignment approaches. 
The top row in Fig.~\ref{fig:toy dataset demo small} corresponds to the domain alignment classifier trained only on the labeled source domain, i.e., $ss$. However, the bottom row in Fig.~\ref{fig:toy dataset demo small} corresponds to contradistinguisher trained using the proposed method \smallmethodname{} with labeled source and unlabeled target domain, i.e., $ss{ + }tu{ + }ta$.

Fig.~\ref{fig:toy dataset plots} demonstrates the classifier learnt using \smallmethodname{} on the synthetic datasets with different complex shapes and orientations of the source and target domain distributions for the input data. 
Fig.~\ref{sfig:cp_src_toy_d0_seed_0022_ss_tu_ta_contour_epoch} and~\ref{sfig:cp_src_toy_d0_seed_3234_ss_tu_ta_contour_epoch} indicate the simplest form of the domain adaptation tasks with similar orientations in source and target domain distributions.
It is important to note that the prior enforcing used in pseudo-label selection is the reason such fine classifier boundaries are observed, especially in Fig.~\ref{sfig:cp_src_toy_d1_seed_0022_ss_tu_ta_contour_epoch},\ref{sfig:cp_src_toy_d1_seed_3234_ss_tu_ta_contour_epoch} and~\ref{sfig:cp_src_toy_d0_seed_1811_ss_tu_ta_contour_epoch}-\ref{sfig:src_toy_d0_seed_2108_ss_tu_ta_contour_epoch006}. 
Fig.~\ref{sfig:src_toy_d0_seed_3182_ss_tu_ta_contour_epoch077} and~\ref{sfig:src_toy_d1_seed_0152_ss_tu_ta_contour_epoch012} represent more complex configurations of source and target domain distributions that indicate the hyperbolic decision boundaries jointly learnt on both the domains simultaneously using a single classifier without explicit domain alignment. Similarly, Fig.~\ref{sfig:src_toy_d1_seed_1887_ss_tu_ta_contour_epoch021} represents a complex configuration of source and target domain distributions that indicates an elliptical decision boundary. 

{\color{black}\color{black}These simulated experiments points to some significant inner workings of our approach \smallmethodname{}. These are the two main takeaways from the toy-dataset experiments.
\noindent
\begin{inparaenum}[(i)]
    \item The non-necessity of the domain alignment in the form of distribution distance metric minimization or data augmentation. In the case of these toy-datasets, it is not possible to perform any form of data augmentation, unlike some of the visual domain adaptation tasks, because the data is directly available in the form of encoded features that cannot be easily data augmented through standard heuristics. In such a case, it is necessary to realize a generic approach applicable to multiple modalities of the input, e.g., similar to the toy-dataset in language domain adaptation tasks. The features are presented in the form of word2vec/doc2vec, and no data augmentation is possible.
    \noindent
    \item Fig.~\ref{sfig:cp_src_toy_d0_seed_3234_ss_contour_epoch} and~\ref{sfig:src_toy_d0_seed_0661_ss_tu_ta_contour_epoch016} provide some interesting observations. Here, we can observe that the classes in the source domain are overlapping, resulting in less than 100\% classification on the source domain, which in turn results in less than 100\% classification on the target domain when considering domain alignment approaches. However, \smallmethodname{} does not try to morph the target domain onto the source domain by directly classifying on the target domain resulting in a perfect classification. Since the classification is done directly on the unlabeled target domain in a fully unsupervised manner, the target domain classification performance is not limited by the source domain classification performance, i.e., the irrespective of the domain is used as the labeled source domain and the unlabeled target domain, the performance is the respective domains are similar. In other words, swapping of the domains or the direction of the domain adaptation has little effect on the classification performance on each individual domain.
\end{inparaenum}
}
\color{black}

\subsection{Experiments on Real-world Datasets} \label{sec:setup}
{
In our previous work~\cite{Balgi2019CUDA}, we have demonstrated the effectiveness of \smallmethodname{} in real-world domain adaptation on low-resolution visual datasets and language datasets.
In contrast to low-resolution visual datasets, we consider the complex, high-resolution Office-31~\cite{DBLP:conf/eccv/SaenkoKFD10} and {VisDA-2017~\cite{peng2017visda}} object classification datasets for domain adaptation. In addition to the single-source domain adaptation experiments, we also extend \smallmethodname{} to Office-31~\cite{DBLP:conf/eccv/SaenkoKFD10} and Digits datasets~\cite{lecun1989backpropagation, lecun1998gradient, 37648, pmlr-v37-ganin15}.
}

\subsubsection{Office-31 Dataset} \label{sec: high res visual exp setup office}
\begin{figure*}[ht!]
\begin{center}
\centering
\includegraphics[width=\linewidth,height=\textheight,keepaspectratio=true]{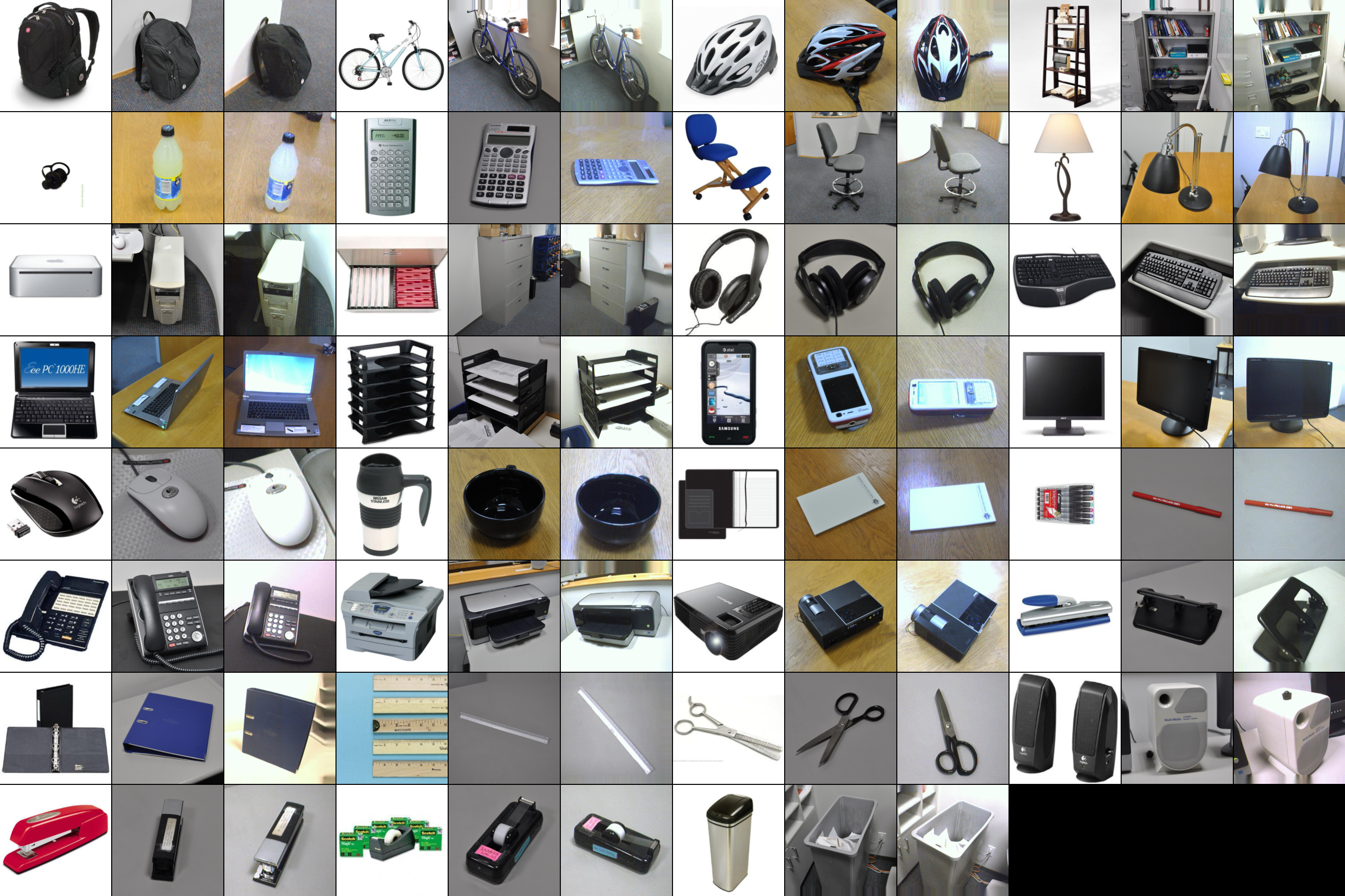}
\caption{Illustrations of samples from all the three domains of high resolution Office-31~\cite{DBLP:conf/eccv/SaenkoKFD10} dataset with one instance per each class from every domain (column \{1,4,7,10\}: $\mathcal{A}$, \{2,5,8,11\}: $\mathcal{D}$, \{3,6,9,12\}: $\mathcal{W}$). (Best viewed in color.)
}
\label{fig: high res visual datasets}
\end{center}
\end{figure*}
\begin{figure*}[ht!]
\begin{center}
\centering
\includegraphics[width=\linewidth,height=\textheight,keepaspectratio=true]{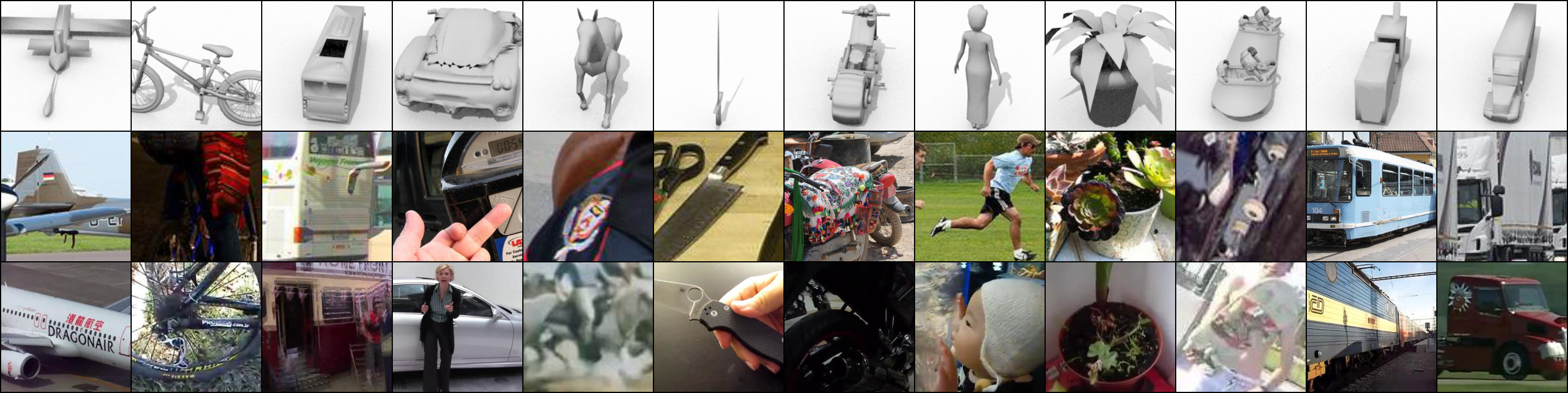}
\caption{{Illustrations of samples from all the three data-splits of VisDA-2017~\cite{peng2017visda} dataset with one instance per each class from every domain (\{row 1\}: $\mathcal{V}_{syn}$ source domain synthetic images (training set), \{row 2\}: $\mathcal{V}_{real}$ target domain real-world images (validation set), \{row 3\}: $\mathcal{V}_{real}$ target domain real-world images (testing set)). {It should be noted that unlike the Office-31 dataset and other standard benchmark domain adaptation datasets discussed in~\cite{Balgi2019CUDA}, most of the real-world images in the target domain of the VisDA-2017 dataset contains multiple true labels, which are only annotated with only one of the multiple labels.} (Best viewed in color.)}
}
\label{fig: visda17 datasets}
\end{center}
\end{figure*}
In high-resolution visual datasets, we consider Office-31~\cite{DBLP:conf/eccv/SaenkoKFD10} dataset for our experiments. Unlike low-resolution visual datasets, here, we have only a few hundreds of training samples that make this an even more challenging task.

\textbf{Office objects}:
\href{https://drive.google.com/open?id=0B4IapRTv9pJ1WGZVd1VDMmhwdlE}{Office-31}~\cite{DBLP:conf/eccv/SaenkoKFD10} dataset consists of high resolution images of objects belonging to 31 classes obtained from three different domains AMAZON~($\mathcal{A}$), DSLR~($\mathcal{D}$) and WEBCAM~($\mathcal{W}$).
Fig.~\ref{fig: high res visual datasets} shows illustrations of the images from all the three above mentioned domains of the Office-31~\cite{DBLP:conf/eccv/SaenkoKFD10} dataset. AMAZON~($\mathcal{A}$) domain consists of synthetic images with clear white background. DSLR~($\mathcal{D}$) and WEBCAM~($\mathcal{W}$) domains consist of real-world images with noisy background and surroundings. We consider all possible six combinatorial tasks of domain adaptation involving all the three domains, i.e., $\mathcal{A}{ \leftrightarrow }\mathcal{D}$, $\mathcal{A}{ \leftrightarrow }\mathcal{W}$ and $\mathcal{D}{ \leftrightarrow }\mathcal{W}$. Compared to low-resolution visual datasets, Office-31~\cite{DBLP:conf/eccv/SaenkoKFD10} dataset domain adaptation tasks have increased complexity due to the small number of training images. 
Unlike low-resolution visual datasets, the high-resolution Office-31~\cite{DBLP:conf/eccv/SaenkoKFD10} dataset does not have separate pre-defined train and test splits. Since we do not use any labels from the target domain during training, we report ten-crop test accuracy on the target domain by summing the softmax values of all the ten crops of the image and assign the label with maximum aggregate softmax value for the given image as in CDAN~\cite{NIPS2018_7436} in Table~\ref{tbl:image results office31}.

To further alleviate the lack of a large number of training samples, pre-trained networks such as ResNet-50~\cite{he2016deep} and ResNet-152~\cite{he2016deep} were used to extract 2048 dimensional features from high-resolution images similar to CDAN~\cite{NIPS2018_7436}. 
Since the images are not well centered and have a high resolution, we use the standard ten-crop of the image to extract features from the same images during training and testing, also similar to CDAN~\cite{NIPS2018_7436}. 

The use of pre-trained models leads to two choices of training, 
\begin{inparaenum}[(i)]
\item Fine-tune the pre-trained model used as feature extractor along with the final classifier layer: This requires careful selection of several hyper-parameters such as learning rate, learning rate decay, batch size, etc., to fine-tune the network to the current dataset while preserving the ability of the pre-trained network. We observed that fine-tuning also depends on the loss function used for training~\cite{DBLP:conf/iclr/JacobsenBZB19}, which in our case, the use of contradistinguish loss greatly affected the changes in the pre-trained model as it is trained only using cross-entropy loss. Fine-tuning is also computationally expensive and time-consuming as each iteration requires computing gradients of all the pre-trained model parameters. 
\item Fix the pre-trained model and only train the final classifier layer: Alternative to fine-tuning is to fix the pre-trained model and use it only as a feature extractor. This approach has multiple practical benefits such as,
\begin{inparaenum}[(a)]
\item The computational time and cost of fine-tuning the parameters of the pre-trained model are alleviated.
\item Since the feature extractor is fixed, it requires only once to extract and store the features locally instead of extracting the same features every iteration. Hence reducing the training time and the GPU memory as it is only required to train the final classifier.
\end{inparaenum}
\end{inparaenum}

\subsubsection{{VisDA-2017 Dataset}} \label{sec: high res visual exp setup visda}
{
The VisDA-2017 dataset consists of two domains, (i) synthetic and (ii) real, with three predefined data splits. Fig.~\ref{fig: visda17 datasets} indicates the samples from all the 12 classes of the three data splits.
The three predefined data splits in the VisDA-2017 dataset are as follows.}

\noindent
\begin{inparaenum}[(i)]
{\item Training set: This split includes 152,397 labeled synthetic images obtained using 2D renderings of 3D models from different angles and different lighting conditions. This split is considered as a labeled source domain for training.}

\noindent
{\item Validation set: This split includes 55,388 real-world images obtained from a curated subset of MS COCO~\cite{DBLP:conf/eccv/LinMBHPRDZ14} dataset. This split is considered an unlabeled target domain training set, and this is used during the training without labels.}

\noindent
{\item Testing set: This split includes 72,372 real-world images obtained from YouTube Bounding Boxes~\cite{DBLP:conf/cvpr/RealSMPV17} dataset. This split is considered as the target domain testing set used for evaluation and to report the results.}
\end{inparaenum}

\subsection{Analysis of Experimental Results on Real-world Datasets} \label{sec:results}
\begin{figure*}[ht!]
\begin{center}
\hfill
\subfloat[$\mathcal{A}{ \rightarrow }\mathcal{D}$]{
\includegraphics[width=0.3\linewidth,height=0.3\textheight,keepaspectratio=true]{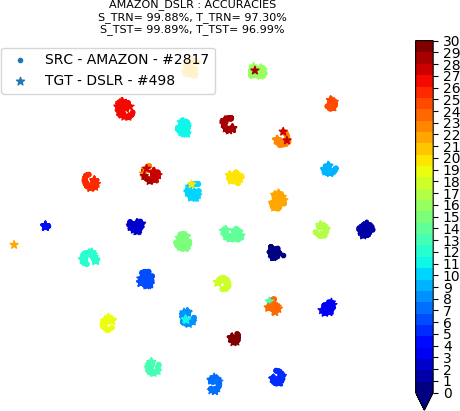}
\label{sfig:tsne a_d152}
}
\hfill
\subfloat[$\mathcal{A}{ \rightarrow }\mathcal{W}$]{
\includegraphics[width=0.3\linewidth,height=0.3\textheight,keepaspectratio=true]{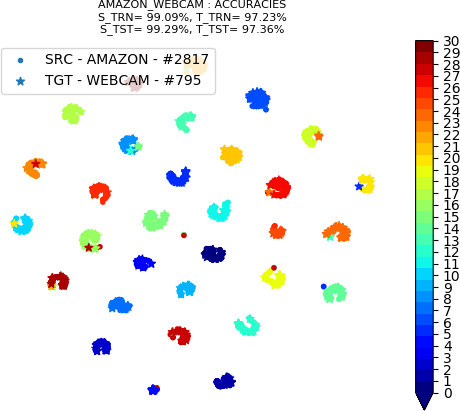}
\label{sfig:tsne a_w152}
}
\hfill
\subfloat[$\mathcal{D}{ \rightarrow }\mathcal{W}$]{
\includegraphics[width=0.3\linewidth,height=0.3\textheight,keepaspectratio=true]{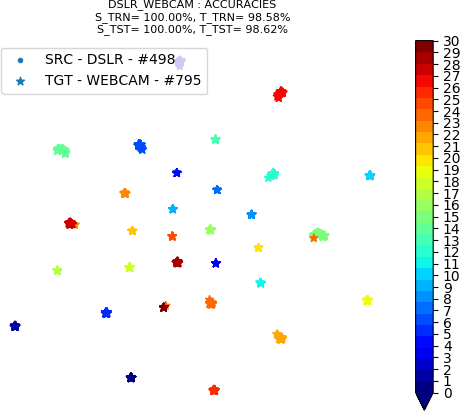}
\label{sfig:tsne d_w152}
}
\hfill
\\
\hfill
\subfloat[$\mathcal{D}{ \rightarrow }\mathcal{A}$]{
\includegraphics[width=0.3\linewidth,height=0.3\textheight,keepaspectratio=true]{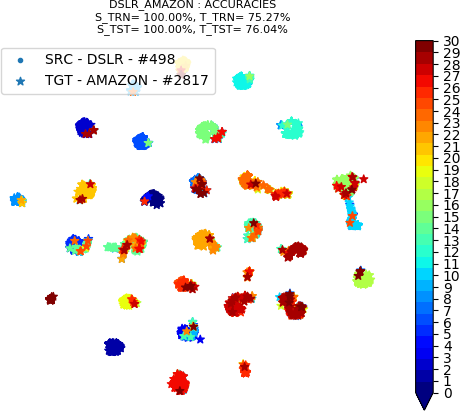}
\label{sfig:tsne d_a152}
}
\hfill
\subfloat[$\mathcal{W}{ \rightarrow }\mathcal{A}$]{
\includegraphics[width=0.3\linewidth,height=0.3\textheight,keepaspectratio=true]{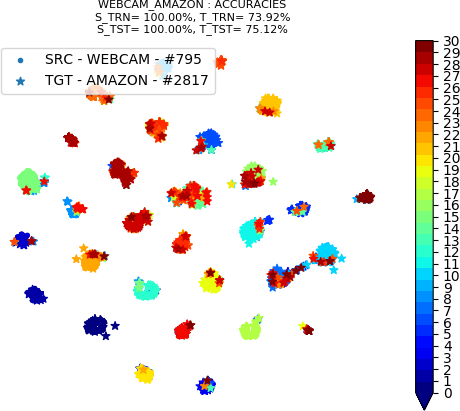}
\label{sfig:tsne w_a152}
}
\hfill
\subfloat[$\mathcal{W}{ \rightarrow }\mathcal{D}$]{
\includegraphics[width=0.3\linewidth,height=0.3\textheight,keepaspectratio=true]{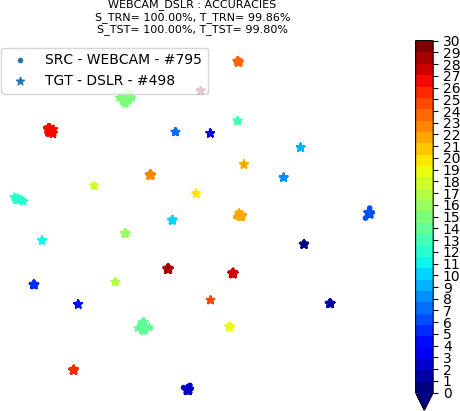}
\label{sfig:tsne w_d152}
}
\hfill
\\
\hfill
\subfloat[$\mathcal{A}{ + }\mathcal{D}{ \rightarrow }\mathcal{W}$]{
\includegraphics[width=0.3\linewidth,height=0.3\textheight,keepaspectratio=true]{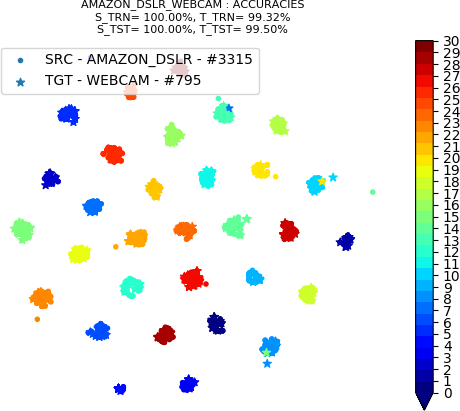}
\label{sfig:tsne a_d_w50}
}
\hfill
\subfloat[$\mathcal{D}{ + }\mathcal{W}{ \rightarrow }\mathcal{A}$]{
\includegraphics[width=0.3\linewidth,height=0.3\textheight,keepaspectratio=true]{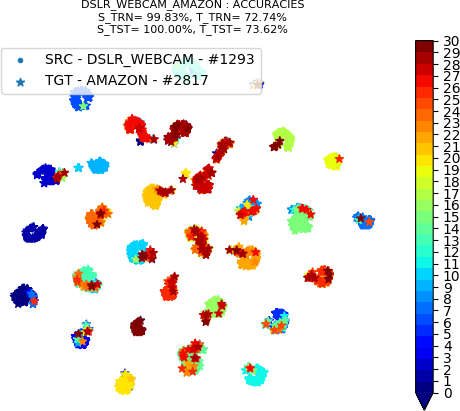}
\label{sfig:tsne d_w_a50}
}
\hfill
\subfloat[$\mathcal{W}{ + }\mathcal{A}{ \rightarrow }\mathcal{D}$]{
\includegraphics[width=0.3\linewidth,height=0.3\textheight,keepaspectratio=true]{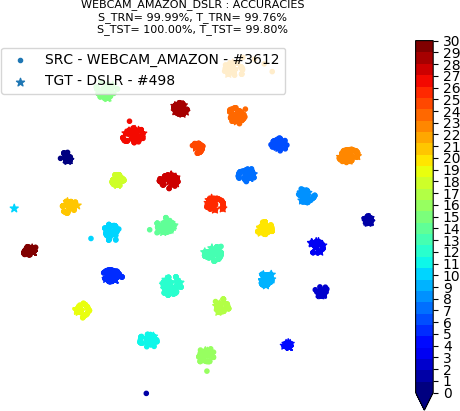}
\label{sfig:tsne w_a_d50}
}
\hfill

\caption{
{Row 1 and 2: t-SNE~\cite{vandermaaten2008visualizing} plots for embeddings from the output of contradistinguisher with samples from Office-31~\cite{DBLP:conf/eccv/SaenkoKFD10} dataset as input corresponding to the highest mean accuracy setting $ss{ + }tu{ + }su{ + }ta$ indicated in Table~\ref{tbl:image results office31} for single-source domain adaptation using ResNet-152~\cite{he2016deep} as the fixed encoder. 
Row 3: t-SNE~\cite{vandermaaten2008visualizing} plots for embeddings from the output of contradistinguisher corresponding to the samples from Office-31~\cite{DBLP:conf/eccv/SaenkoKFD10} dataset in high-resolution visual tasks after applying softmax trained with \smallmethodname{} with ResNet-50~\cite{he2016deep} as the encoder in a \textbf{multi-source domain adaptation} setting as indicated in Table~\ref{tbl: office31 multisource}.
We can observe the clear class-wise clustering among all the 31 classes in the Office-31~\cite{DBLP:conf/eccv/SaenkoKFD10} datasets. 
We achieve high accuracies in spite of having only a few hundred training samples in each domain.
(Best viewed in color.)
}
}
\label{fig:tsne office31 res50_152}
\end{center}
\end{figure*}
\begin{table*}[ht!]
\centering
\caption{Target domain accuracy (\%) on high resolution Office-31~\cite{DBLP:conf/eccv/SaenkoKFD10} dataset containing three domains. 
\smallmethodname{} corresponds to our best results obtained with the best hyper-parameter settings. 
$ss$: source supervised~\eqref{eq:ce}, $tu$: target unsupervised~\eqref{eq:unsup}, $su$: source unsupervised~\eqref{eq:unsup}, $sa$: source adversarial regularization~\eqref{eq:adv reg loss} and $ta$: target adversarial regularization~\eqref{eq:adv reg loss} represents different training configurations.
}
\label{tbl:image results office31}
 \setlength\tabcolsep{6pt}
\begin{tabular}{lccccccc}
\toprule
\textbf{Method} & $\mathcal{A}{ \rightarrow }\mathcal{D}$ & $\mathcal{A}{ \rightarrow }\mathcal{W}$ & $\mathcal{D}{ \rightarrow }\mathcal{A}$ & $\mathcal{D}{ \rightarrow }\mathcal{W}$ & $\mathcal{W}{ \rightarrow }\mathcal{A}$ & $\mathcal{W}{ \rightarrow }\mathcal{D}$ & $Mean$ \\
\midrule
DAN~\cite{DBLP:conf/icml/LongC0J15} & 78.6 & 80.5 & 63.6 & 97.1 & 62.8 & 99.6 & 80.3 \\ 
RTN~\cite{DBLP:conf/nips/LongZ0J16} & 77.5 & 84.5 & 66.2 & 96.8 & 64.8 & 99.4 & 81.5 \\ 
JAN~\cite{DBLP:conf/icml/LongZ0J17} & 84.7 & 85.4 & 68.6 & 97.4 & 70.0 & 99.8 & 84.3 \\ 
Rozantsev et. al.~\cite{rozantsev2018beyond} & 75.5 & 75.8 & 55.7 & 96.7 & 57.6 & 99.6 & 76.8 \\ 
{CAN~\cite{DBLP:conf/cvpr/Kang0YH19}} & {95.0} & {94.5} & {\textbf{78.0}} & {99.1} & {\textbf{77.0}} & {99.8} & \textbf{90.6} \\ 
RevGrad~\cite{pmlr-v37-ganin15} & 79.7 & 82.0 & 68.2 & 96.9 & 67.4 & 99.1 & 82.2 \\ 
ADDA~\cite{8099799} & 77.8 & 86.2 & 69.5 & 96.2 & 68.9 & 98.4 & 82.8 \\
GTA~\cite{DBLP:conf/cvpr/Sankaranarayanan18a} & 87.7 & 89.5 & 72.8 & 97.9 & 71.4 & 99.8 & 86.5 \\ 
CDAN~\cite{NIPS2018_7436} & 92.9 & 94.1 & 71.0 & 98.6 & 69.3 & \textbf{100.0} & 87.6 \\ 
{DICE}~\cite{liang2018aggregating} & 68.5 & 72.5 & 58.1 & 97.2 & 60.3 & \textbf{100.0} & 76.1 \\ 
{ATM~\cite{9080115}} & \textbf{96.4} & 95.7 & 74.1 & \textbf{99.3} & 73.5 & \textbf{100.0} & 89.8 \\ 
{CADA-P~\cite{DBLP:conf/cvpr/KurmiKN19}} & 95.6 & \textbf{97.0} & 71.5 & \textbf{99.3} & 73.1 & \textbf{100.0} & 89.5 \\ 
{PFAN~\cite{DBLP:conf/cvpr/ChenXHRD0XH19}} & 76.3 & 83.0 & 63.3 & 99.0 & 60.8 & 99.9 & 80.4 \\ 
\cmidrule{2-8}
\color{black}\color{black}\textbf{\smallmethodname{}} (Ours) (with ResNet-50) & 96.0 & 95.6 & 73.2 & 99.1 & 74.7 & \textbf{100.0} & 89.8\color{black} \\
\midrule
\textbf{\smallmethodname{}} (Ours) (with ResNet-152) & \textbf{97.0} & \textbf{98.5} & 76.0 & 99.0 & 76.0 & \textbf{100.0} & \textbf{91.1} \\
\midrule
$ss$ (Ours) (fine-tune ResNet-50) & 41.0 & 38.7 & 23.2 & 80.6 & 25.6 & 94.2 & 50.6 \\
$ss$ (Ours) (fixed ResNet-50) & 82.0 & 77.9 & 68.4 & 97.2 & 67.1 & \textbf{100.0} & 82.1 \\
$ss{ + }tu$ (Ours) (fixed ResNet-50) & 95.0 & 93.8 & 71.5 & 98.9 & 73.3 & 99.4 & 88.7 \\
$ss{ + }tu{ + }su$ (Ours) (fixed ResNet-50) & \textbf{96.0} & \textbf{95.6} & 69.5 & \textbf{99.1} & 70.7 & \textbf{100.0} & 88.5 \\
$ss{ + }tu{ + }su{ + }ta$ (Ours) (fixed ResNet-50) & 92.8 & 91.6 & 72.5 & 98.4 & 72.8 & 99.8 & 88.0 \\
$ss{ + }tu{ + }su{ + }ta{ + }sa$ (Ours) (fixed ResNet-50) & 91.8 & \textbf{95.6} & \textbf{73.2} & 98.0 & \textbf{74.7} & \textbf{100.0} & \textbf{88.9} \\
\cmidrule{2-8}
$ss$ (Ours) (fixed ResNet-152) & 84.9 & 82.8 & 70.3 & 98.2 & 71.1 & \textbf{100.0} & 84.6 \\
$ss{ + }tu$ (Ours) (fixed ResNet-152) & \textbf{97.0} & 94.3 & 73.9 & \textbf{99.0} & 75.5 & \textbf{100.0} & 90.0 \\
$ss{ + }tu{ + }su$ (Ours) (fixed ResNet-152) & 95.6 & 95.6 & 73.8 & 98.7 & 74.3 & \textbf{100.0} & 89.7 \\
$ss{ + }tu{ + }su{ + }ta$ (Ours) (fixed ResNet-152) & \textbf{97.0} & 97.4 & \textbf{76.0} & 98.6 & 75.1 & 99.8 & \textbf{90.7} \\
$ss{ + }tu{ + }su{ + }ta{ + }sa$ (Ours) (fixed ResNet-152) & 95.4 & \textbf{98.5} & 75.0 & 98.9 & \textbf{76.0} & \textbf{100.0} & 90.6 \\
\bottomrule
\end{tabular}
\end{table*}
\begin{table*}[ht!]
    \centering
    \caption{Target domain accuracy (\%) on high resolution Office-31~\cite{DBLP:conf/eccv/SaenkoKFD10} dataset under \textbf{multi-source domain adaptation} setting by combining two domains into a single source domain and the remaining domain as the target domain with ResNet-50~\cite{he2016deep} as the encoder. 
\smallmethodname{} corresponds to our best results obtained with the best hyper-parameter settings. 
$ss$: source supervised~\eqref{eq:ce}, $tu$: target unsupervised~\eqref{eq:unsup}, $su$: source unsupervised~\eqref{eq:unsup}, $sa$: source adversarial regularization~\eqref{eq:adv reg loss} and $ta$: target adversarial regularization~\eqref{eq:adv reg loss} represents different training configurations.
}
    \label{tbl: office31 multisource}
\begin{tabular}{llcccc}
\toprule
Setting & Method & $\mathcal{A}{ + }\mathcal{D}{ \rightarrow }\mathcal{W}$ & $\mathcal{D}{ + }\mathcal{W}{ \rightarrow }\mathcal{A}$ & $\mathcal{W}{ + }\mathcal{A}{ \rightarrow }\mathcal{D}$ & $Mean$ \\
\midrule
\multirow{11}{*}{Best single source} & DAN~\cite{DBLP:conf/icml/LongC0J15} & 97.1 & 63.6 & 99.6 &  86.8 \\
 & RTN~\cite{DBLP:conf/nips/LongZ0J16} & 96.8 & 66.2 & 99.4 & 87.5 \\ 
& JAN~\cite{DBLP:conf/icml/LongZ0J17} & 97.4 & 70.0 & 99.8 & 89.1 \\ 
& Rozantsev et. al.~\cite{rozantsev2018beyond} & 96.7 & 57.6 & 99.6 & 84.6 \\ 
& mDA-layer~\cite{8792192} & 94.5 & 64.9 & 94.9 & 84.8 \\ 
& RevGrad~\cite{pmlr-v37-ganin15} & 96.9 & 68.2 & 99.1 & 88.1 \\ 
& ADDA~\cite{8099799} & 96.2 & 69.5 & 98.4 & 88.0 \\
& GTA~\cite{DBLP:conf/cvpr/Sankaranarayanan18a} & 97.9 & 72.8 & 99.8 & 90.2 \\ 
& CDAN~\cite{NIPS2018_7436} & 98.6 & 71.0 & \textbf{100.0} & 89.8 \\ 
& {DICE}~\cite{liang2018aggregating} & 97.2 & 60.3 & \textbf{100.0} & 85.8 \\ 
\cmidrule{2-6}
& \textbf{\smallmethodname{}} (Ours) & \textbf{99.1} & \textbf{74.7} & \textbf{100.0} & \textbf{91.3} \\
\midrule
\multirow{13}{*}{Multi-source} & DAN~\cite{DBLP:conf/icml/LongC0J15} & 95.2 & 53.4 & 98.8 &  82.5 \\
 & mDA-layer~\cite{8792192} & 94.6 & 62.6 & 93.7 & 83.6 \\ 
 & DIAL~\cite{carlucci2017just} & 94.3 & 62.5 & 93.8 & 83.5 \\
 & SGF~\cite{gopalan2011domain} & 52.0 & 28.0 & 39.0 & 39.7 \\
 & sFRAME~\cite{xie2015learning} & 52.2 & 32.1 & 54.5 & 46.3 \\
 & RevGrad~\cite{pmlr-v37-ganin15} & 96.2 & 54.6 & 98.8 & 83.2 \\ 
 & DCTN~\cite{xu2018deep} & 96.9 & 54.9 & 99.6 & 83.8 \\
\cmidrule{2-6}
& \textbf{\smallmethodname{}} (Ours) & \textbf{99.5} & \textbf{73.6} & \textbf{99.8} & \textbf{91.0} \\
\cmidrule{2-6}
& $ss$ (Ours) & 95.6 & 68.1 & 99.2 & 87.6 \\
& $ss{ + }tu$ (Ours) & 99.4 & 72.1 & \textbf{99.8} & 90.4 \\
& $ss{ + }tu{ + }su$ (Ours) & 98.9 & 70.3 & 99.4 & 89.5 \\
& $ss{ + }tu{ + }su{ + }ta$ (Ours) & \textbf{99.5} & 73.3 & 99.6 & 90.8 \\
& $ss{ + }tu{ + }su{ + }ta{ + }sa$ (Ours) & 99.4 & \textbf{73.6} & 99.2 & 90.7 \\
 \bottomrule
\end{tabular}
\end{table*}
\begin{table*}[ht!]
\centering
\caption{\color{black}\color{black}Target domain accuracy reported on the test set (\%) on all 5 combinations of Digits datasets under \textbf{multi-source domain adaptation} setting.\color{black}
}
\label{tbl:image results digits}
 \setlength\tabcolsep{6pt}
\begin{tabular}{lcccccc}
\toprule
\multirow{2}{*}{\textbf{Method}} & $mn,sn,sv,us$ & $mm,sn,sv,us$ & $mm,mn,sv,us$ & $mm,mn,sn,us$ & $mm,mn,sn,sv$ & \multirow{2}{*}{$Mean$} \\
 & $\rightarrow mm$ & $\rightarrow mn$ & $\rightarrow sn$ & $\rightarrow sv$ & $\rightarrow us$ &  \\
\midrule
DAN~\cite{DBLP:conf/icml/LongC0J15} & 63.78 & 96.31 & 85.43 & 62.45 & 94.24 & 80.44 \\
M3SDA~\cite{DBLP:conf/iccv/PengBXHSW19} & 72.82 & 98.43 & 89.58 & 81.32 & 96.14 & 87.66 \\
RevGrad/DANN~\cite{pmlr-v37-ganin15} & 71.30 & 97.60 & 85.34 & 63.48 & 92.33 & 82.01 \\
ADDA~\cite{8099799} & 71.57 & 97.89 & 86.45 & 75.48 & 92.83 & 84.84 \\
DCTN~\cite{xu2018deep} & 70.53 & 96.23 & 86.77 & 77.61 & 82.81 & 82.79 \\
\cmidrule{2-7}
\textbf{\smallmethodname{}} (Ours) & \textbf{87.96} & \textbf{99.51} & \textbf{97.96} & \textbf{89.03} & \textbf{99.30} & \textbf{94.75} \\
\bottomrule
\end{tabular}
\end{table*}
\begin{figure*}[ht!]
\begin{center}
\subfloat[{$mn,sn,sv,us\rightarrow mm$}]{
{\includegraphics[width=0.185\linewidth,height=0.185\textheight,keepaspectratio=true]{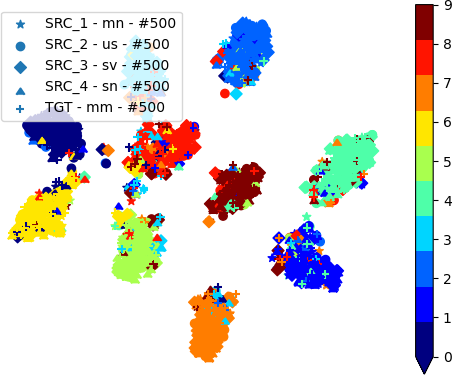}}
\label{sfig:mnistm_target}
}
\hfill
\subfloat[{$mm,sn,sv,us\rightarrow mn$}]{
{\includegraphics[width=0.185\linewidth,height=0.185\textheight,keepaspectratio=true]{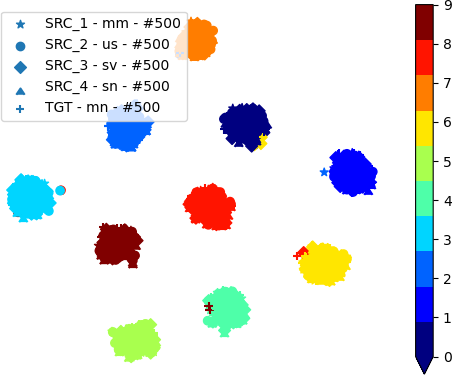}}
\label{sfig:mnist_target}
}
\hfill
\subfloat[{$mm,mn,sv,us\rightarrow sn$}]{
{\includegraphics[width=0.185\linewidth,height=0.185\textheight,keepaspectratio=true]{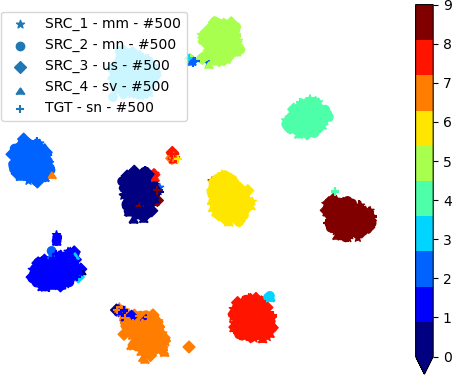}}
\label{sfig:syn_target}
}
\hfill
\subfloat[{$mm,mn,sn,us\rightarrow sv$}]{
{\includegraphics[width=0.185\linewidth,height=0.185\textheight,keepaspectratio=true]{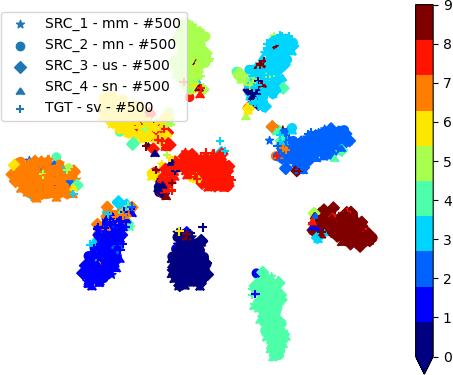}}
\label{sfig:svhn_target}
}
\hfill
\subfloat[{$mm,mn,sn,sv\rightarrow us$}]{
{\includegraphics[width=0.185\linewidth,height=0.185\textheight,keepaspectratio=true]{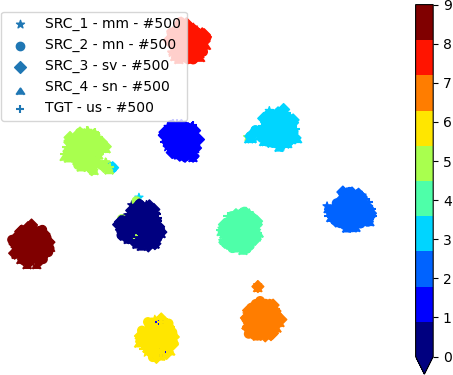}}
\label{sfig:usps_target}
}
\caption{\color{black}\color{black}{
The t-SNE plots of the unseen test set samples corresponding to the \smallmethodname{} result in Table~\ref{tbl:image results digits}. The t-SNE plots show clear clustering of all the 10 classes in Digits datasets distinctively.
(Best viewed in color.)}\color{black}
}
\label{fig:digits multisource tsne plots}
\end{center}
\end{figure*}
\begin{figure*}[ht!]
\begin{center}
\subfloat[{BSP}]{
{\includegraphics[width=0.24\linewidth,height=0.24\textheight,keepaspectratio=true]{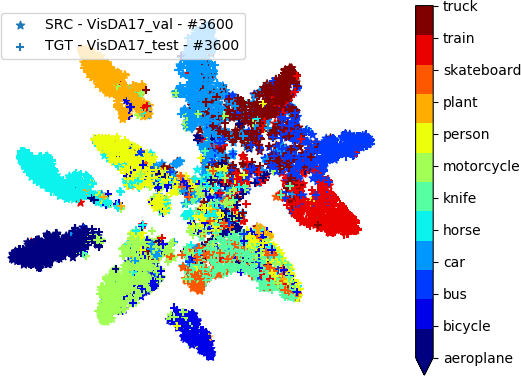}}
\label{sfig:visda17 tsne bsp}
}
\hfill
\subfloat[{CAN}]{
{\includegraphics[width=0.24\linewidth,height=0.24\textheight,keepaspectratio=true]{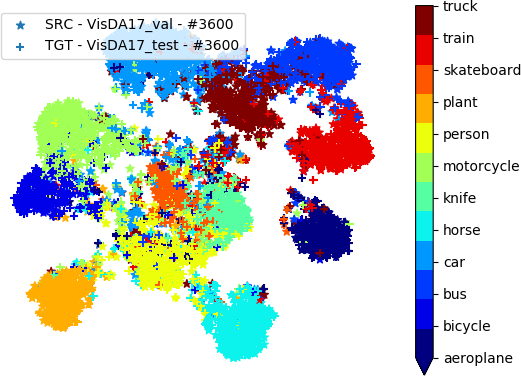}}
\label{sfig:visda17 tsne can}
}
\hfill
\subfloat[{\smallmethodname{}}]{
{\includegraphics[width=0.24\linewidth,height=0.24\textheight,keepaspectratio=true]{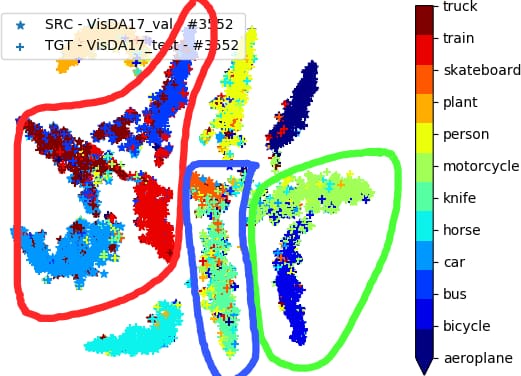}}
\label{sfig:visda17 tsne cuda e5}
}
\hfill
\subfloat[{\smallmethodname{}$^{*}$}]{
{\includegraphics[width=0.24\linewidth,height=0.24\textheight,keepaspectratio=true]{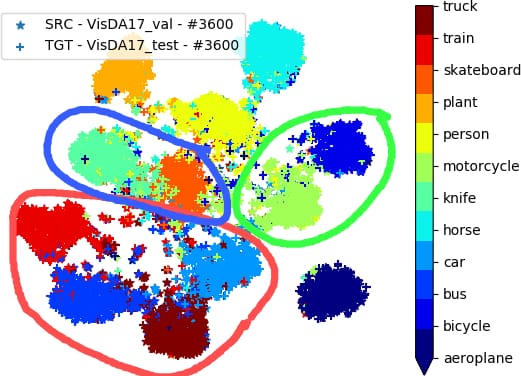}}
\label{sfig:visda17 tsne cuda*}
}
\caption{{
The t-SNE plots of \smallmethodname{}/\smallmethodname{}$^{*}$ shows the clear clustering of all the twelve classes of VisDA-2017 distinctively compared to the t-SNE plots of BSP/CAN. 
The t-SNE plots of \smallmethodname{}/\smallmethodname{}$^{*}$ represent some important visual semantics of the image embeddings obtained from contradistinguisher in the following manner.
(i) The vehicular classes such as `bus', `car', `train', and `truck' can be seen clustered closely as semantically these classes are similar to each other (region bounded in red).
(ii) The two-wheeler classes such as `bicycle' and `motorcycle' are clustered closely as these are semantically similar to each other compared to vehicular classes that are clustered exactly opposite (region bounded in green).
(iii) Irrespective of the approach used, there is always confusion between `knife' and `skateboard' classes. This confusion between `knife' and `skateboard' classes represented in the confusion matrices, which is also clearly seen in the t-SNE plots as well, can be attributed to the nature of images of these classes in the dataset on close observation (region bounded in blue).
(iv) The remaining classes such as `aeroplane', `horse', `person' and `plant' can be seen clustered independently and distinctively as these classes have almost no visual semantic similarities to one another.  
(Best viewed in color.)}
}
\label{fig:visda17 results plots}
\end{center}
\end{figure*}
\begin{table*}[ht!]
    \centering
    \caption{{Results on VisDA-2017 dataset reproduced from the current state-of-the-art method BSP, CAN and our proposed method~\smallmethodname{} reported on both the validation set and test set. We report all the evaluation metrics such as precision, recall, and accuracy, unlike BSP/CAN, where the recall scores are mistakenly reported as accuracy. CUDA$^{*}$ represents the results reproduced using vanilla CUDA with the data augmentation and target domain clustering similar to CAN for a fair comparison of the effect of the CUDA over CAN.
    The results reported for BSP, CAN, and CUDA/CUDA$^{*}$ are from our own best reproduction from the original source code.
    }}
    \label{tbl: visda results}
\resizebox{\linewidth}{!}{%
\begin{tabular}{lllccccccccccccc}
\toprule
Data-split & Metric (\%) & Method & Aeroplane & Bicycle & Bus & Car & Horse & Knife & Motorcycle & Person & Plant & Skateboard & Train & Truck & $Mean$ \\
\midrule
\multirow{13}{*}{Validation} & \multirow{4}{*}{Precision} & BSP~\cite{chen2019transferability} & 93.79 & 91.40 & 71.03 & 78.36 & 94.98 & 35.25 & 78.90 & 63.12 & 91.37 & 55.28 & 82.16 & 58.15 & 74.48 \\
 & & {CAN~\cite{DBLP:conf/cvpr/Kang0YH19}} & 96.92 & 85.87 & 83.45 & 85.06 & 92.18 & 70.73 & 91.25 & 61.38 & 91.77 & 63.45 & 89.91 & 74.94 & 82.24 \\
 &  & CUDA (Ours) & 96.33 & 87.04 & 79.20 & 80.68 & 92.25 & 67.66 & 86.74 & 70.28 & 91.57 & 67.37 & 87.36 & 50.18 & 79.72 \\
 &  & {CUDA$^{*}$ (Ours)} & 97.08 & 86.27 & 83.71 & 86.03 & 91.56 & 77.53 & 91.59 & 67.23 & 95.47 &
 64.66 & 90.89 & 72.89 & \textbf{83.74} \\
\cmidrule{2-16}
 & \multirow{4}{*}{Recall} & BSP~\cite{chen2019transferability} & 90.35 & 71.54 & 76.42 & 59.11 & 87.91 & 80.58 & 89.41 & 73.75 & 84.70 & 76.41 & 81.35 & 45.84 & 76.45 \\
 & & {CAN~\cite{DBLP:conf/cvpr/Kang0YH19}} & 96.60  & 88.63 & 81.90  & 68.47 & 96.25 & 95.95 & 87.80  & 80.45 & 96.88 & 94.74 & 85.01 & 60.04 & 86.06 \\
  &  & CUDA (Ours) & 91.47 & 83.11 & 75.59 & 69.16 & 94.65 & 93.35 & 88.13 & 82.52 & 90.79 & 81.02 & 82.22 & 51.69 & 81.97  \\
  &  & {CUDA$^{*}$ (Ours)} & 96.76 & 90.04 & 83.94 & 70.82 & 96.18 & 96.63 & 88.53 & 86.28 & 95.43 &
 94.48 & 85.98 & 62.09 & \textbf{87.26}  \\
\cmidrule{2-16}
& \multirow{5}{*}{Accuracy} & TPN~\cite{DBLP:conf/cvpr/PanYLWNM19} & 93.70 & 85.10 & 69.20 & 81.60 & 93.50 & 61.90 & 89.30 & 81.40 & 93.50 & 81.60 & 84.50 & 49.90 & 80.40 \\
 \cmidrule{3-16}
 & & BSP~\cite{chen2019transferability} & 98.97 & 97.79 & 95.36 & 89.26 & 98.58 & 93.73 & 96.39 & 94.99 & 98.09 & 96.48 & 97.22 & 91.27 & 95.68 \\
 & & {CAN~\cite{DBLP:conf/cvpr/Kang0YH19}} & 99.57 & 98.37 & 97.09 & 91.82 & 98.99 & 98.36 & 97.84 & 94.93 & 99.03 & 97.54 & 98.12 & 93.99 & 97.14 \\
   &  & CUDA (Ours) & 99.21 & 98.16 & 96.25 & 91.10 & 98.87 & 98.08 & 97.35 & 96.22 & 98.56 & 97.60 & 97.73 & 90.02 & 96.60   \\
  &  & {CUDA$^{*}$ (Ours)} & 99.60  & 98.48 & 97.26 & 92.36 & 98.93 & 98.82 & 97.95 & 95.97 & 99.25 &
 97.65 & 98.27 & 93.89 & \textbf{97.37}   \\
\midrule
\multirow{12}{*}{Test} & \multirow{4}{*}{Precision} & BSP~\cite{chen2019transferability} & 91.76 & 95.83 & 71.50 & 83.08 & 94.87 & 33.08 & 76.45 & 72.25 & 87.38 & 34.54 & 83.36 & 80.12 & 75.35 \\
 & & {CAN~\cite{DBLP:conf/cvpr/Kang0YH19}} & 95.72 & 89.20  & 86.38 & 93.31 & 95.28 & 74.33 & 91.12 & 76.74 & 86.90  & 45.56 & 93.92 & 87.24 & 84.64 \\
  &  & CUDA (Ours) & 93.31 & 94.44 & 81.28 & 82.26 & 91.77 & 66.12 & 80.43 & 77.37 & 88.48 & 45.19 & 89.43 & 60.74 & 79.24 \\
  &  & {CUDA$^{*}$ (Ours)} & 95.18 & 87.61 & 86.79 & 92.01 & 96.16 & 77.36 & 90.60  & 81.28 & 93.49 &
 47.33 & 94.15 & 85.83 & \textbf{85.65} \\
\cmidrule{2-16}
 & \multirow{4}{*}{Recall} & BSP~\cite{chen2019transferability} & 77.98 & 44.64 & 79.44 & 91.11 & 72.75 & 70.26 & 73.99 & 54.85 & 88.69 & 44.50 & 73.84 & 67.84 & 69.99 \\
 & & {CAN~\cite{DBLP:conf/cvpr/Kang0YH19}} & 92.01 & 74.79 & 85.60  & 91.84 & 87.33 & 89.26 & 68.09 & 80.89 & 97.34 & 84.79 & 81.94 & 89.24 & 85.26\\
  &  & CUDA (Ours) & 82.41 & 58.45 & 71.41 & 92.16 & 83.70 & 79.58 & 72.46 & 75.55 & 93.03 & 70.78 & 73.95 & 79.58 & 77.75 \\
  &  & {CUDA$^{*}$ (Ours)} & 92.44 & 80.97 & 86.41 & 92.18 & 87.85 & 89.42 & 69.35 & 84.20  & 96.41 &
 85.34 & 83.48 & 89.27 & \textbf{86.44} \\
\cmidrule{2-16}
 & \multirow{4}{*}{Accuracy} & BSP~\cite{chen2019transferability} & 97.92 & 96.62 & 94.97 & 97.24 & 97.27 & 86.96 & 94.55 & 92.98 & 98.57 & 94.66 & 95.89 & 95.42 & 95.25  \\
 & & {CAN~\cite{DBLP:conf/cvpr/Kang0YH19}} & 99.13 & 97.98 & 97.31 & 98.52 & 98.51 & 96.85 & 95.70  & 95.38 & 98.97 & 95.55 & 97.66 & 97.77 & 97.44\\
  &  & CUDA (Ours) & 98.31 & 97.34 & 95.66 & 97.21 & 97.92 & 95.36 & 94.96 & 95.06 & 98.87 & 95.61 & 96.51 & 93.28 & 96.34 \\
  &  & {CUDA$^{*}$ (Ours)} & 99.12 & 98.20  & 97.42 & 98.41 & 98.63 & 97.21 & 95.78 & 96.27 & 99.39 &
 95.82 & 97.82 & 97.62 & \textbf{97.64} \\
 \bottomrule
\end{tabular}
}
\end{table*}
\begin{table}[ht!]
    \centering
    \caption{{Total classification accuracy (\%) on VisDA-2017 dataset reported on both the validation set and test set. The results from JAN, GTA, CDAN and TransNorm are reported from TransNorm~\cite{wang2019transferable}. 
    The results reported for BSP, CAN and CUDA/CUDA$^{*}$ are from our own best reproduction from the original source code.
     CUDA$^{*}$ represents the results reproduced using vanilla CUDA with the data augmentation and target domain clustering similar to CAN for a fair comparison of the effect of the CUDA over CAN.}}
    \label{tbl: visda results total_acc}
\begin{tabular}{lcc}
\toprule
    \multirow{2}{*}{Method} & \multicolumn{2}{c}{Data-split}\\
    \cmidrule{2-3}
 & Validation & Test \\
\midrule
JAN~\cite{DBLP:conf/icml/LongZ0J17} &  61.6 & - \\
GTA~\cite{DBLP:conf/cvpr/Sankaranarayanan18a} & 69.5 & - \\
CDAN~\cite{NIPS2018_7436} & 70.0 & - \\
TransNorm~\cite{wang2019transferable} & 71.4 & - \\
BSP~\cite{chen2019transferability} & 74.1 & 71.5 \\
{CAN~\cite{DBLP:conf/cvpr/Kang0YH19}} & {82.8} & {84.7} \\
\smallmethodname{} (Ours) & 79.6 & 78.1 \\
{\smallmethodname{}$^{*}$ (Ours)} & {\textbf{84.2}} & {\textbf{85.8}} \\
 \bottomrule
\end{tabular}
\end{table}
\renewcommand{\thesubfigure}{\arabic{subfigure}}
\begin{figure*}[ht!]
\begin{center}
\subfloat[aeroplane|person]{
\includegraphics[width=0.156\linewidth,height=0.156\textheight,keepaspectratio=true]{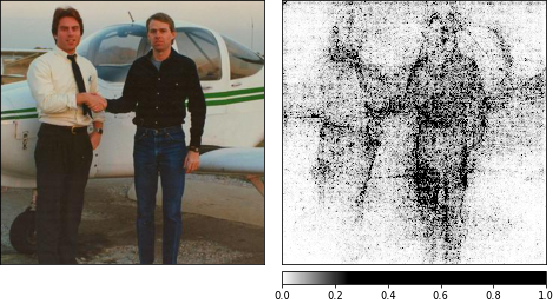}
\label{sfig:visda17 misclass noise_tunnel_1_aeroplane_003130_person}
}
\hfill
\subfloat[aeroplane|train]{
\includegraphics[width=0.156\linewidth,height=0.156\textheight,keepaspectratio=true]{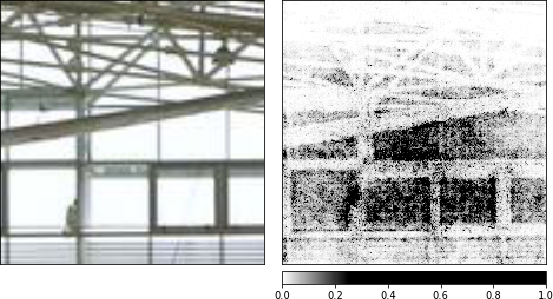}
\label{sfig:visda17 misclass noise_tunnel_1_aeroplane_003126_train}
}
\hfill
\subfloat[aeroplane|truck]{
\includegraphics[width=0.156\linewidth,height=0.156\textheight,keepaspectratio=true]{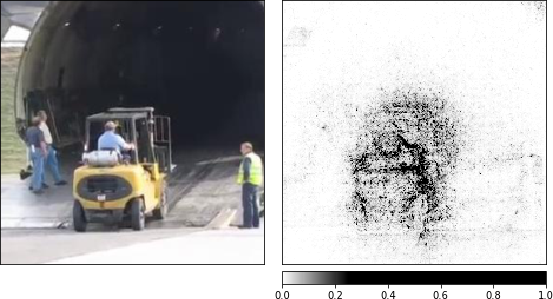}
\label{sfig:visda17 misclass noise_tunnel_2_aeroplane_062320_truck}
}
\hfill
\subfloat[bicycle|motorcycle]{
\includegraphics[width=0.156\linewidth,height=0.156\textheight,keepaspectratio=true]{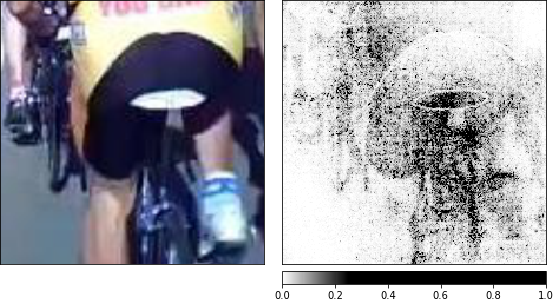}
\label{sfig:visda17 misclass noise_tunnel_2_bicycle_046644_motorcycle}
}
\hfill
\subfloat[bus|train]{
\includegraphics[width=0.156\linewidth,height=0.156\textheight,keepaspectratio=true]{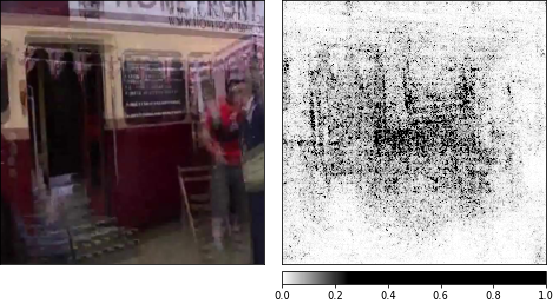}
\label{sfig:visda17 misclass noise_tunnel_2_bus_train}
}
\hfill
\subfloat[bus|car]{
\includegraphics[width=0.156\linewidth,height=0.156\textheight,keepaspectratio=true]{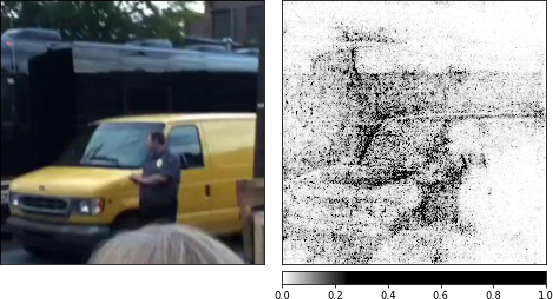}
\label{sfig:visda17 misclass noise_tunnel_2_bus_041588_car}
}
\\
\subfloat[bus|person]{
\includegraphics[width=0.156\linewidth,height=0.156\textheight,keepaspectratio=true]{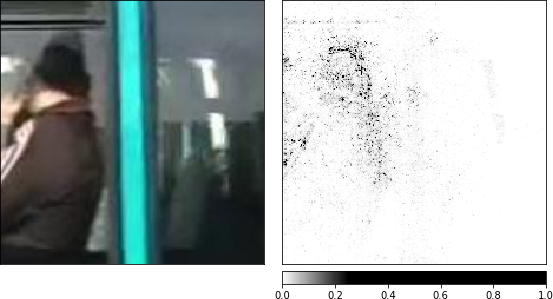}
\label{sfig:visda17 misclass noise_tunnel_2_bus_068364_person}
}
\hfill
\subfloat[bus|train]{
\includegraphics[width=0.156\linewidth,height=0.156\textheight,keepaspectratio=true]{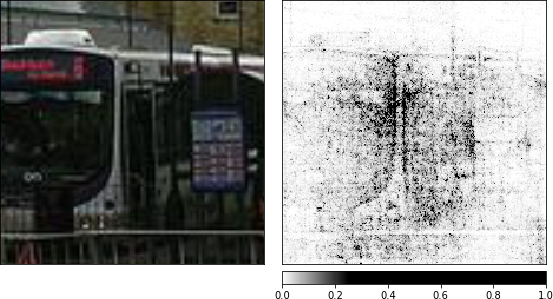}
\label{sfig:visda17 misclass noise_tunnel_1_bus_010017_train}
}
\hfill
\subfloat[bus|truck]{
\includegraphics[width=0.156\linewidth,height=0.156\textheight,keepaspectratio=true]{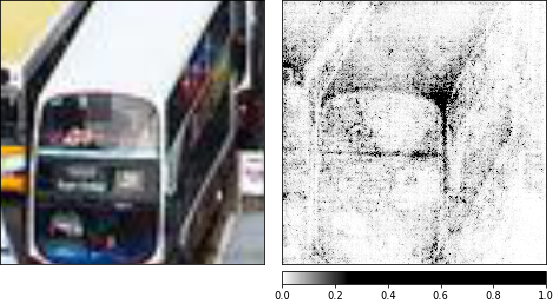}
\label{sfig:visda17 misclass noise_tunnel_1_bus_011195_truck}
}
\hfill
\subfloat[car|person]{
\includegraphics[width=0.156\linewidth,height=0.156\textheight,keepaspectratio=true]{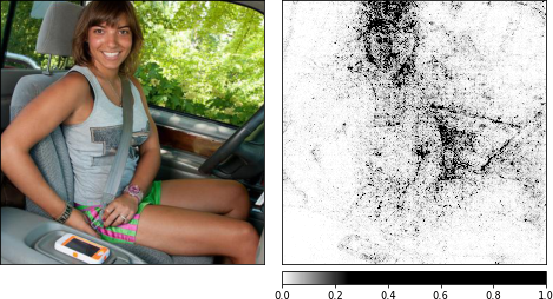}
\label{sfig:visda17 misclass noise_tunnel_1_car_017315_person}
}
\hfill
\subfloat[horse|person]{
\includegraphics[width=0.156\linewidth,height=0.156\textheight,keepaspectratio=true]{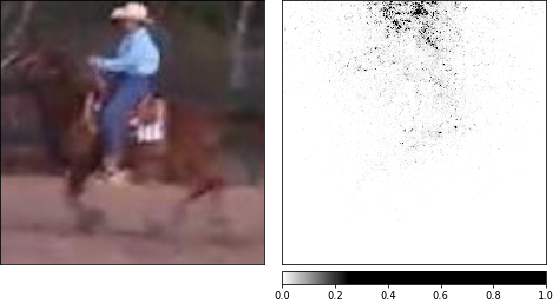}
\label{sfig:visda17 misclass noise_tunnel_2_horse_051900_person}
}
\hfill
\subfloat[horse|person]{
\includegraphics[width=0.156\linewidth,height=0.156\textheight,keepaspectratio=true]{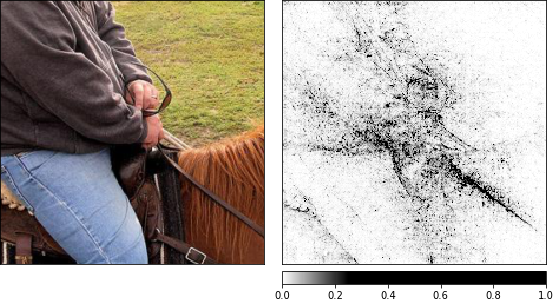}
\label{sfig:visda17 misclass noise_tunnel_1_horse_023721_person}
}
\\
\subfloat[{motorcycle\newline}|{truck}]{
\includegraphics[width=0.156\linewidth,height=0.156\textheight,keepaspectratio=true]{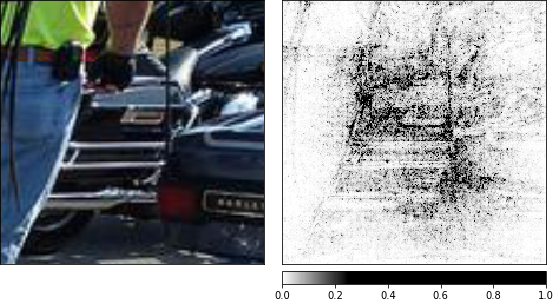}
\label{sfig:visda17 misclass noise_tunnel_1_motorcycle_032635_truck}
}
\hfill
\subfloat[{motorcycle\newline}|{plant}]{
\includegraphics[width=0.156\linewidth,height=0.156\textheight,keepaspectratio=true]{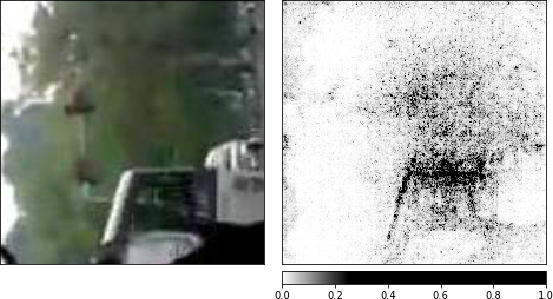}
\label{sfig:visda17 misclass noise_tunnel_2_motorcycle_034841_plant}
}
\hfill
\subfloat[person|horse]{
\includegraphics[width=0.156\linewidth,height=0.156\textheight,keepaspectratio=true]{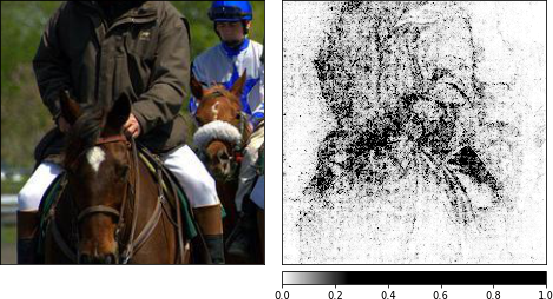}
\label{sfig:visda17 misclass noise_tunnel_1_person_38335_horse}
}
\hfill
\subfloat[{person\newline}|{motorcycle}]{
\includegraphics[width=0.156\linewidth,height=0.156\textheight,keepaspectratio=true]{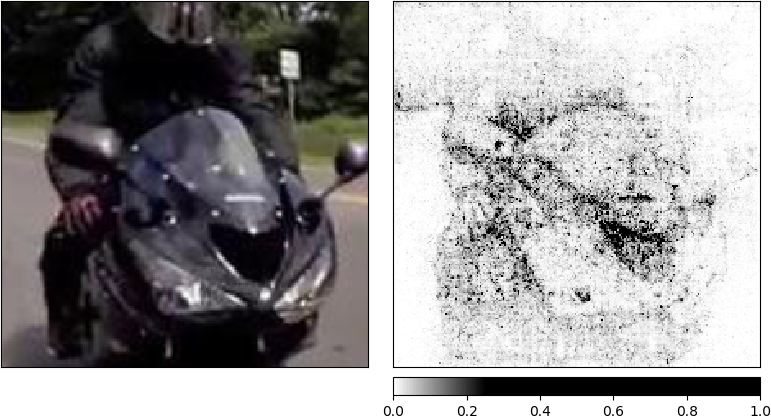}
\label{sfig:visda17 misclass noise_tunnel_2_person_040272_motorcycle}
}
\hfill
\subfloat[plant|person]{
\includegraphics[width=0.156\linewidth,height=0.156\textheight,keepaspectratio=true]{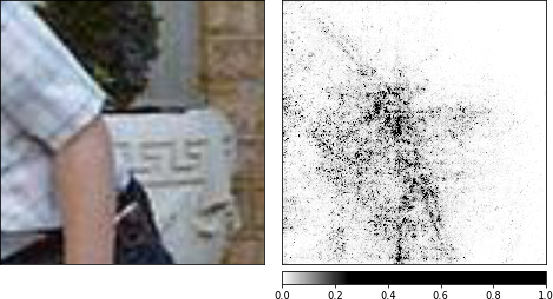}
\label{sfig:visda17 misclass noise_tunnel_1_plant_042539_person}
}
\hfill
\subfloat[{skateboard\newline}|{person}]{
\includegraphics[width=0.156\linewidth,height=0.156\textheight,keepaspectratio=true]{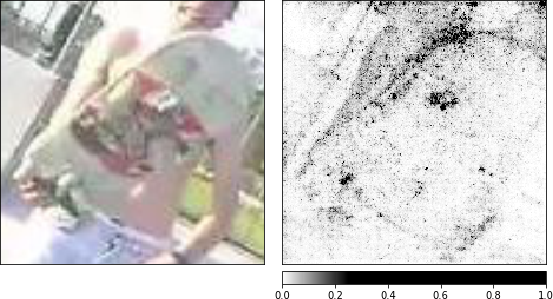}
\label{sfig:visda17 misclass noise_tunnel_2_skateboard_person}
}
\\
\subfloat[train|bus]{
\includegraphics[width=0.156\linewidth,height=0.156\textheight,keepaspectratio=true]{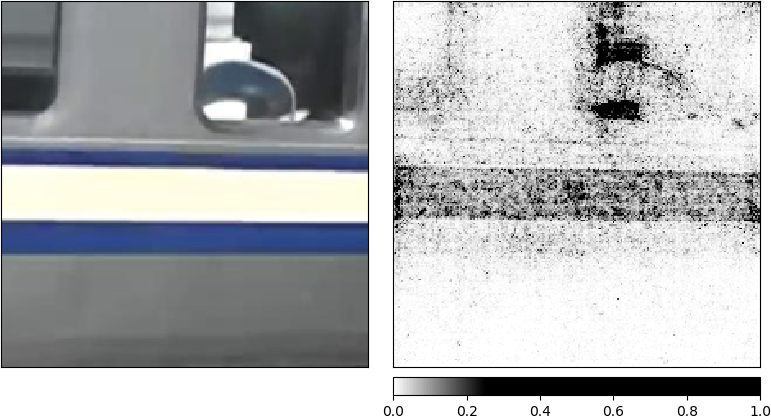}
\label{sfig:visda17 misclass noise_tunnel_2_train_020016_bus}
}
\hfill
\subfloat[train|bus]{
\includegraphics[width=0.156\linewidth,height=0.156\textheight,keepaspectratio=true]{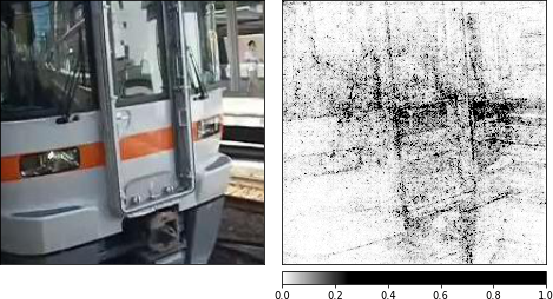}
\label{sfig:visda17 misclass noise_tunnel_2_train_041888_bus}
}
\hfill
\subfloat[train|bus]{
\includegraphics[width=0.156\linewidth,height=0.156\textheight,keepaspectratio=true]{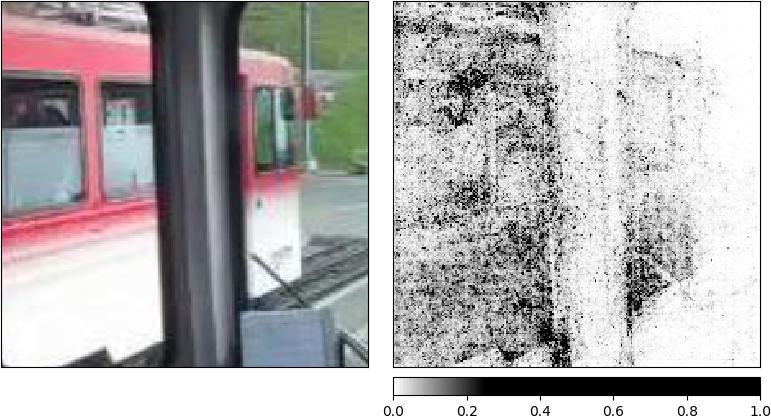}
\label{sfig:visda17 misclass noise_tunnel_2_train_047964_bus}
}
\hfill
\subfloat[train|bus]{
\includegraphics[width=0.156\linewidth,height=0.156\textheight,keepaspectratio=true]{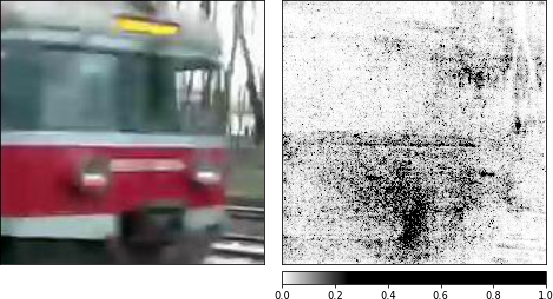}
\label{sfig:visda17 misclass noise_tunnel_2_train_041288_bus}
}
\hfill
\subfloat[train|person]{
\includegraphics[width=0.156\linewidth,height=0.156\textheight,keepaspectratio=true]{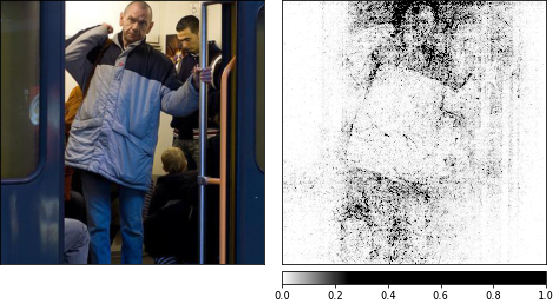}
\label{sfig:visda17 misclass noise_tunnel_1_train_046874_person}
}
\hfill
\subfloat[train|person]{
\includegraphics[width=0.156\linewidth,height=0.156\textheight,keepaspectratio=true]{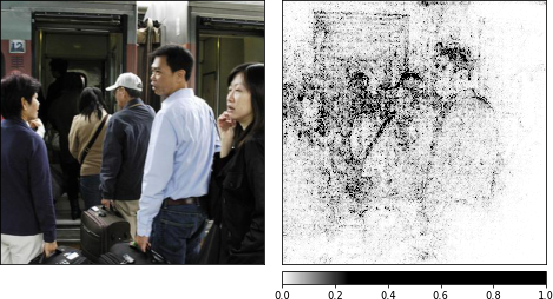}
\label{sfig:visda17 misclass noise_tunnel_1_train_048539_person}
}
\\
\subfloat[train|person]{
\includegraphics[width=0.156\linewidth,height=0.156\textheight,keepaspectratio=true]{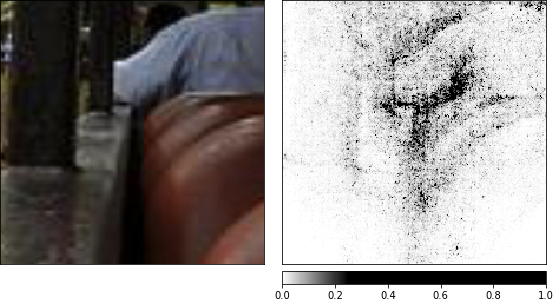}
\label{sfig:visda17 misclass noise_tunnel_1_train_049102_person}
}
\hfill
\subfloat[train|knife]{
\includegraphics[width=0.156\linewidth,height=0.156\textheight,keepaspectratio=true]{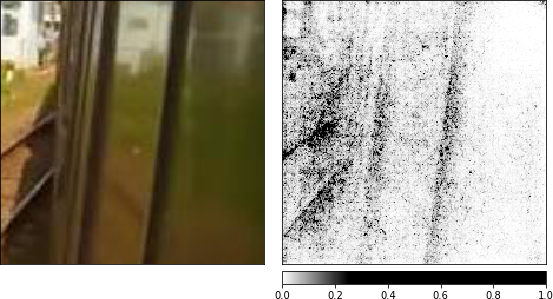}
\label{sfig:visda17 misclass noise_tunnel_2_train_004178_knife}
}
\hfill
\subfloat[train|truck]{
\includegraphics[width=0.156\linewidth,height=0.156\textheight,keepaspectratio=true]{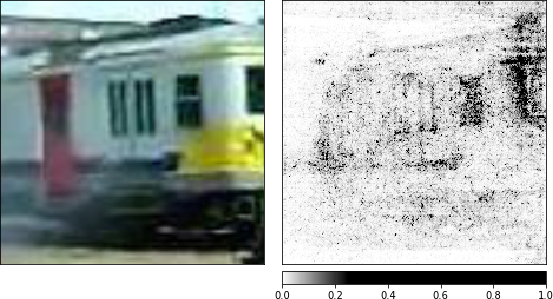}
\label{sfig:visda17 misclass noise_tunnel_2_train_054601_truck}
}
\hfill
\subfloat[truck|bus]{
\includegraphics[width=0.156\linewidth,height=0.156\textheight,keepaspectratio=true]{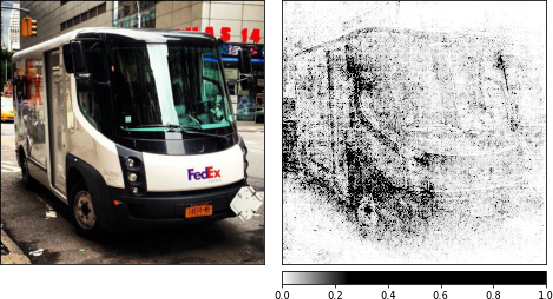}
\label{sfig:visda17 misclass noise_tunnel_1_truck_049903_bus}
}
\hfill
\subfloat[truck|bus]{
\includegraphics[width=0.156\linewidth,height=0.156\textheight,keepaspectratio=true]{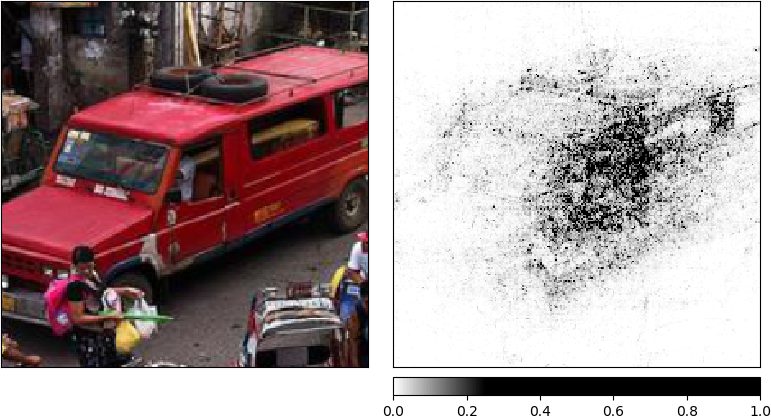}
\label{sfig:visda17 misclass noise_tunnel_1_truck_054176_bus}
}
\subfloat[truck|bus]{
\includegraphics[width=0.156\linewidth,height=0.156\textheight,keepaspectratio=true]{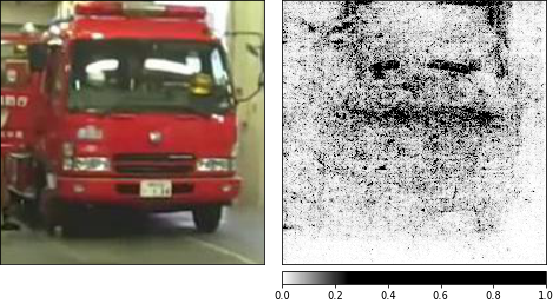}
\label{sfig:visda17 misclass noise_tunnel_2_truck_022023_bus}
}
\\
\subfloat[truck|bus]{
\includegraphics[width=0.156\linewidth,height=0.156\textheight,keepaspectratio=true]{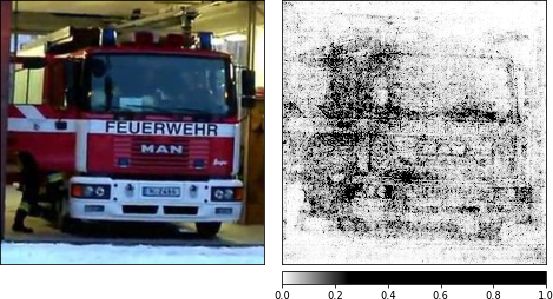}
\label{sfig:visda17 misclass noise_tunnel_2_truck_021831_bus}
}
\hfill
\subfloat[truck|bus]{
\includegraphics[width=0.156\linewidth,height=0.156\textheight,keepaspectratio=true]{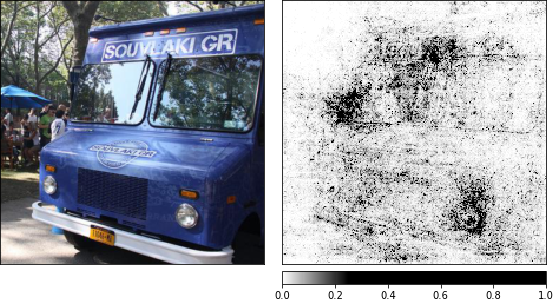}
\label{sfig:visda17 misclass noise_tunnel_1_truck_053664_bus}
}
\hfill
\subfloat[truck|car]{
\includegraphics[width=0.156\linewidth,height=0.156\textheight,keepaspectratio=true]{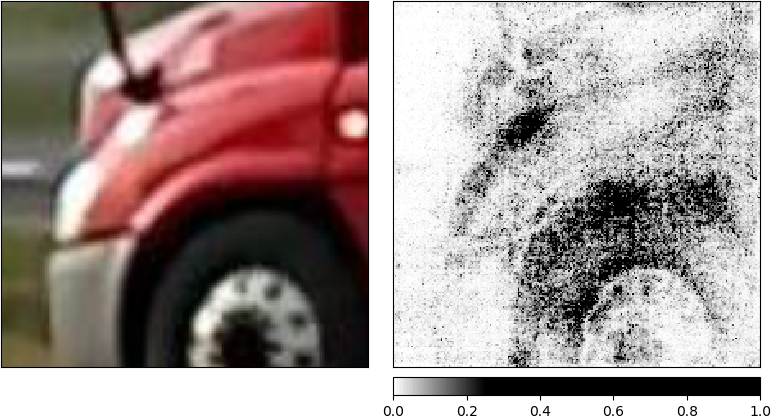}
\label{sfig:visda17 misclass noise_tunnel_2_truck_036870_car}
}
\hfill
\subfloat[truck|car]{
\includegraphics[width=0.156\linewidth,height=0.156\textheight,keepaspectratio=true]{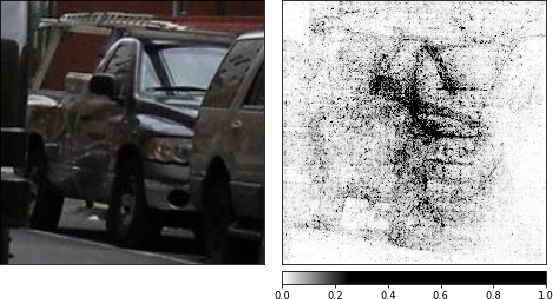}
\label{sfig:visda17 misclass noise_tunnel_1_truck_050874_car}
}
\hfill
\subfloat[truck|car]{
\includegraphics[width=0.156\linewidth,height=0.156\textheight,keepaspectratio=true]{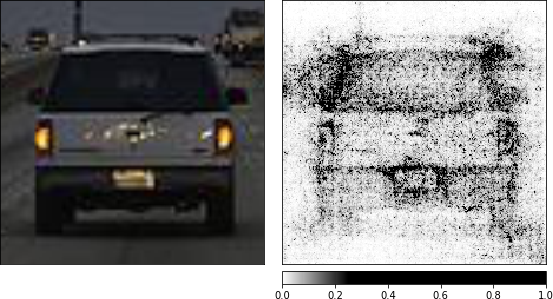}
\label{sfig:visda17 misclass noise_tunnel_1_truck_052191_car}
}
\hfill
\subfloat[truck|person]{
\includegraphics[width=0.156\linewidth,height=0.156\textheight,keepaspectratio=true]{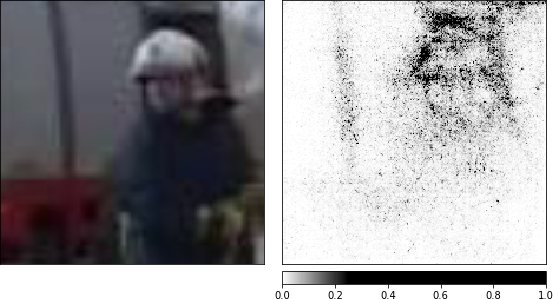}
\label{sfig:visda17 misclass noise_tunnel_2_truck_037630_person}
}
\caption{{We indicate few samples that are misclassified by the contradistinguisher in the following subcaption format `original\_label|predicted\_label'. 
In most cases, the original ground truth labels are dubious, and the predicted labels make more sense realistically. 
{Subplots (6), (7), (11), (12), (14), (18), (23) and (25) shows that the object is identified based on the shape and not if the object is present only in the foreground. This indicates that the contradistinguisher makes the predictions based on the clearly visible shapes and not the presence of the object in the foreground/background.}
The visualization of the features responsible for the respective predicted outcome indicates the shape bias as mostly the features are detected as edges corresponding to the shape of the object in the image. This shows the importance of shape bias to achieve high performance in transfer learning and domain adaptation tasks.}}
\label{fig:visda17 misclassification plots}
\end{center}
\end{figure*}
\renewcommand{\thesubfigure}{\alph{subfigure}}
\subsubsection{Office-31 Single-Source Domain Adaptation Results} \label{sec: high res visual exp analysis}
We report the standard ten-crop accuracy on the target domain images as reported by several state-of-the-art domain adaptation methods~\cite{NIPS2018_7436, DBLP:conf/cvpr/Sankaranarayanan18a, DBLP:conf/icml/LongZ0J17}. Since there are no explicit test split specified in the dataset and no labels are used from the target domain during training, it is common to report ten-crop accuracy considering the whole target domain. 

In Table~\ref{tbl:image results office31}, we report accuracies obtained by fine-tuning ResNet-50~\cite{he2016deep} using the learning rate schedule followed in CDAN~\cite{NIPS2018_7436} and also without fine-tuning ResNet-50~\cite{he2016deep}. 
Apart from fixed ResNet-50~\cite{he2016deep}, we also report accuracies with fixed ResNet-152~\cite{he2016deep} in Table~\ref{tbl:image results office31} for comparison. Fig.~\ref{sfig:tsne a_d152}-\ref{sfig:tsne w_d152} indicate the t-SNE~\cite{vandermaaten2008visualizing} plots of the softmax output after aggregating the ten-crop of each image corresponding to training configuration $ss{ + }tu{ + }su{ + }ta$ reported in Table~\ref{tbl:image results office31}. Fig.~\ref{fig:tsne office31 res50_152} reports the t-SNE~\cite{vandermaaten2008visualizing} plots of the training setting using ResNet-152~\cite{he2016deep} encoder with the highest mean accuracy of all the six domain adaptation tasks. We clearly observe that~\smallmethodname{} outperforms several state-of-the-art methods that also use ResNet-50~\cite{he2016deep} and even further surpasses by using ResNet-152~\cite{he2016deep} encoder with~\smallmethodname{}. 

Among the three domains in Office-31~\cite{DBLP:conf/eccv/SaenkoKFD10} dataset, $\mathcal{A}$ can be considered as a well-curated synthetic dataset with clear background and $\{\mathcal{D},\mathcal{W}\}$ as an uncurated real-world dataset with noisy background and surroundings. 
We report the six domain adaptation tasks in the order of their complexity from low to high as,
\begin{inparaenum}[(i)]
\item Fig.~\ref{sfig:tsne d_w152} and~\ref{sfig:tsne w_d152} indicate highest accuracies because of similar real-world to real-world domain adaptation task, 
\item Fig.~\ref{sfig:tsne a_d152} and~\ref{sfig:tsne a_w152} indicate moderately high accuracies because of synthetic to real-world domain adaptation task and 
\item Fig.~\ref{sfig:tsne d_a152} and~\ref{sfig:tsne w_a152} indicate the lowest accuracies among all the six tasks because of real-world to synthetic domain adaptation task. 
\end{inparaenum}
Comparing \smallmethodname{} with $ss$ in Tables~\ref{tbl:image results office31} and~\ref{tbl: office31 multisource}, we can see significant improvements in the target domain test accuracies due to the use of contradistinguish loss~\eqref{eq:unsup} demonstrating the effectiveness of contradistinguisher.
~As our method is mainly dependent on the contradistinguish loss~\eqref{eq:unsup},
we observed further improved results by experimenting with better neural networks, e.g., using ResNet-152 over ResNet-50 along with our contradistinguish loss~\eqref{eq:unsup}.
{From our ablations study in Table~\ref{tbl:image results office31}, we observe the effect of selection of ImageNet pre-trained ResNet-50 and ResNet-152 models on the domain adaptation with similar implications with the work~\cite{DBLP:conf/wacv/Zhang020}.
In general, one can always obtain better results irrespective of the approach by using better/deeper pre-trained models and/or data augmentation. 
However, since our main aim is to isolate, observe and benchmark only the true effect of different benchmark approaches, in our experiments, we maintain all the other parameters such as pre-trained neural network/data augmentation similar except for the core idea of the approaches and report our results both on Office-31 and VisDA-2017 datasets.
}



\subsubsection{Multi-Source Domain Adaptation Results} \label{sec: high res visual exp analysis multisource}
We also extend the experiments to multi-source domain adaptation on the Office-31~\cite{DBLP:conf/eccv/SaenkoKFD10} and Digits datasets~\cite{lecun1989backpropagation, lecun1998gradient, 37648, pmlr-v37-ganin15}. 
In Table~\ref{tbl: office31 multisource}, we can clearly observe that in  $\mathcal{A}{ + }\mathcal{D}{ \rightarrow }\mathcal{W}$ task, multi-source domain adaptation provides better results than their respective best single source domain adaptation experiments. However in case of $\mathcal{D}{ + }\mathcal{W}{ \rightarrow }\mathcal{A}$ and $\mathcal{W}{ + }\mathcal{A}{ \rightarrow }\mathcal{D}$, the multi-source domain adaptation improves over $ss$, it underperforms compared to best single source domain adaptation task. This can be attributed to the fact that the model tends to overfit on the source domains resulting in a negative transfer.
This negative transfer behavior is also prevalent in other multi-source domain adaptation approaches since all the other multi-source domain adaptation methods also underperform compared to their best single source domain adaptation results, as reported in Table~\ref{tbl: office31 multisource}. Fig.~\ref{sfig:tsne a_d_w50}-\ref{sfig:tsne w_a_d50} indicates t-SNE~\cite{vandermaaten2008visualizing} plots for embeddings from the output of contradistinguisher corresponding to the samples from Office-31~\cite{DBLP:conf/eccv/SaenkoKFD10} dataset after applying softmax trained with \smallmethodname{} with ResNet-50~\cite{he2016deep} as the encoder in a \textbf{multi-source domain adaptation} setting. We can observe the best results when the target domain is one of the real-world domain, i.e., $\mathcal{D}$ and $\mathcal{W}$. It was consistently observed that domain adaptation tasks with synthetic domain $\mathcal{A}$ as the target domain to be the most complex tasks of all the domain adaptation tasks across all the domain adaptation methods. 

\color{black}\color{black}Similarly, Table~\ref{tbl:image results digits} presents the results of multi-source domain adaptation on Digits datasets against benchmark approaches. In Fig.~\ref{fig:digits multisource tsne plots}, we see the t-SNE plots on the test set depicting clear class-wise clustering that indicates the efficacy of \smallmethodname{} single-source to multi-source extension.
\color{black}








\subsubsection{{VisDA-2017 Single-Source Domain Adaptation Results}} \label{sec: high res visual exp visda17 analysis}
{For experiments on the VisDA-2017 dataset, we consider the most recent state-of-the-art benchmark domain adaptation approaches BSP~\cite{chen2019transferability} and CAN~\cite{DBLP:conf/cvpr/Kang0YH19}. Like BSP and CAN, we use the same neural network architecture with Imagenet pre-trained ResNet-101 with contradistinguish loss for training. BSP and CAN report the evaluation metric of accuracy in their papers. However, on reproducing the results with BSP\footnote{\url{https://github.com/thuml/Batch-Spectral-Penalization}} and CAN\footnote{{\url{https://github.com/kgl-prml/Contrastive-Adaptation-Network\\-for-Unsupervised-Domain-Adaptation}}} to set the baseline for comparison, we noticed that the results reported had the following inconsistencies.}

\begin{inparaenum}[(i)]
    {\item 
The results reported in the paper as accuracy in actual were the class-wise \emph{recall} scores.}

\noindent
{\item The most standard procedure in machine learning is to report experimental performances on 
the test split, which is unseen during the training. However, in BSP and CAN, the results are reported on the validation set, which is used as the unlabeled target domain training set. Since the VisDA-2017 dataset has pre-defined splits for evaluation, reporting the results on the validation set used during the training does not indicate these models' generalizing capability, which is the most important aspect one would base an evaluation.}
\end{inparaenum}
{
We correct the above misreporting by reproducing the results from BSP and CAN and report all the relevant metrics for both validation and test splits of the VisDA-2017 dataset.
Apart from these, we also validate our results on the official challenge evaluation portal}\footnote{{\url{https://competitions.codalab.org/competitions/17052\#results}}} 
{
by submitting the results of our approach CUDA on the VisDA-2017 dataset.
}

{In Fig.~\ref{fig:visda17 results plots}, we report the t-SNE plots reproduced from the current state-of-the-art unsupervised domain adaptation approach BSP~\cite{chen2019transferability} and  CAN~\cite{DBLP:conf/cvpr/Kang0YH19}, in comparison with our approach~\smallmethodname{}/\smallmethodname{}$^*$ on both the pre-defined validation split (seen target domain training set) and testing split (unseen target domain testing set) of the VisDA-2017 dataset. 
The results reported as CUDA corresponds to the CUDA experiments without any data augmentation using the BSP source code as the baseline to keep all the parameters the same for a fair comparison with BSP.
Similarly, the results reported as CUDA$^*$ corresponds to the CUDA experiments with data augmentation and clustering of high confidence target domain samples using the CAN source code as the baseline to keep all the parameters same for a fair comparison with CAN.
We can see the classwise clusters in BSP/CUDA are narrower compared to CAN/CUDA$^*$, which are broader due to the use of data augmentation. Data augmentation helps modify/broaden the data distribution aiding in an improvement over the vanilla approaches without data augmentation.
We further validate the results best results from our approach, i.e., CUDA$^*$ by submitting to the official challenge evaluation leaderboard.
In Table~\ref{tbl: visda results} we compare the per-class precision, recall, and accuracies on both the pre-defined validation set and test set of the VisDA-2017 against the results reproduced from the BSP/CAN. 
In Table~\ref{tbl: visda results total_acc} we compare the total classification accuracies on both the pre-defined validation set and  test set of the VisDA-2017 dataset against different benchmark methods. 
The results in Tables~\ref{tbl: visda results} and~\ref{tbl: visda results total_acc} indicate the superior performance of the proposed method \smallmethodname{}/CUDA$^*$ over the current state-of-the-art domain alignment approaches BSP/CAN on both the pre-defined validation set and a test set of VisDA-2017 dataset. 
Even though the validation set and test set belong to the real-world domain, there is an inherent domain shift between them as both the data splits are collected from two different datasets, i.e., MS COCO~\cite{DBLP:conf/eccv/LinMBHPRDZ14} and YouTube Bounding Boxes~,\cite{DBLP:conf/cvpr/RealSMPV17} respectively. Results from~\smallmethodname{} indicate a better generalization to real-world domain as the scores in validation and test sets are closer compared to other approaches on the VisDA-2017 dataset.
We can also observe that the t-SNE plot of~\smallmethodname{} in Fig.~\ref{sfig:visda17 tsne cuda e5} and~\ref{sfig:visda17 tsne cuda*} clearly shows the visual semantics captured between the classes of images in the VisDA-2017 dataset.}

{Apart from setting~\smallmethodname{} as the solid baseline for VisDA-2017, we further put a conscious effort to carefully investigate the reasons for the misclassification using the contradistinguisher to check if we can further improve the results.
In most misclassified cases, we have observed that the labels predicted by \smallmethodname{} appeared to be correct in comparison to the ground truth label of the dataset. In Fig.~\ref{fig:visda17 misclassification plots}, we present some of these instances where the predicted label is more close to the real label than the ground truth.
We can explain this misclassification as a limitation of the VisDA-2017 dataset in the following way.
The misclassification observed in Fig.~\ref{fig:visda17 misclassification plots} is due to the fact that the images in the VisDA-2017 dataset consist of objects belonging to more than one of the twelve classes, i.e., the images in the dataset consists of multiple labels for a single image, but the dataset only records one of these several true labels. 
As the assumed task for domain adaptation is single-label multi-class We see this as a limitation of the VisDA-2017 dataset compared other benchmark domain adaptation datasets such as Office-31 or the low-resolution visual datasets demonstrated in our conference paper~\cite{Balgi2019CUDA}. It is necessary that the datasets be consistent in the sense that each image has a unique label corresponding to it so that during the evaluation, there is no ambiguity between the original label and the predicted label from the trained model.
The presence of this ambiguity in the dataset classification would then lead to observing the true evaluation metrices resulting in improper benchmarking for any given approach.
However, in the case of the VisDA-2017 dataset, the predicted label from the model cannot be considered as a wrong label as it contains the object of the predicted label in the image.
We believe that this is one of the reasons for the overall low performance apart from the complexity of the VisDA-2017 dataset compared to other visual datasets.
}
{It should also be observed that CUDA identifies the most distinguishing/prominent and assigns the label irrespective of the position (foreground/background) of the object as indicated in some of the subplots in Fig.~\ref{fig:visda17 misclassification plots}.} 
{These misclassified cases indicates one of the strong drawback/limitation of VisDA-2017 dataset compared to other visual datasets, i.e., VisDA-2017 dataset has image samples with multiple true labels instead of a unique label for each image sample. Since the images might contain multiple true classes for an image, ideally all these true labels are to be associated with the image in the dataset to rightly evaluate any trained model for its efficacy. Because we perform single-label multi-class classification, predicting any one of the true labels of the image should be considered as right for the evaluation metric. However, this is not the case as the dataset does not record all the true labels for the images. So, if one plans to rightly use this dataset, all the true labels are to be annotated for each of the images in the dataset or use other benchmark datasets such as DomainNet/LSDAC (Large Scale Domain Adaptation Challenge) dataset~\cite{DBLP:conf/iccv/PengBXHSW19} that alleviates the problem of multi-labels of VisDA-2017 dataset as DomainNet dataset only consist of single true label per each image in the dataset, resulting in correct evaluation without the issue of misclassification we have indicated above for the VisDA-2017 dataset.}

{Apart from analyzing the limitation of the VisDA-2017 dataset, we also analyze the nature of the feature representations learnt by contradistinguisher.}
{In order to visualize the features that prompted the predicted label, we use Captum\footnote{\url{https://captum.ai/tutorials/Resnet_TorchVision_Interpret}}, an open-source, extensible library for model interpretability built on PyTorch\footnote{\url{https://pytorch.org/}}~\cite{paszke2017automatic}. 
We use gradient-based attribution to compute the integrated gradients for a given image using the predicted label. We obtain the high-level features or the saliency maps~\cite{DBLP:journals/corr/SimonyanVZ13} for the given image.
In terms of high-level features in a given image, one can imagine features such as shape, color, texture, size, etc. to be the features that help in predicting the classifier outcome. 
Out of all these features, the most natural and basic feature influencing the outcome is observed to be the shapes of the objects. 
Extensive research materials in psychology such as~\cite{landau1992syntactic, diesendruck2003specific, potrzeba2015investigating, ritter2017cognitive, hosseini2018assessing
} have indicated that human babies and adults tend to utilize shapes than color/material/texture to assign a word label to the given object. This particular phenomenon is widely termed as `shape bias' in the literature.
However, recently, it was shown that the ImageNet pre-trained models possess a texture bias over shape bias~\cite{DBLP:conf/iclr/GeirhosRMBWB19}. To improve the shape bias,~\cite{DBLP:conf/iclr/GeirhosRMBWB19} propose a new modified dataset called `Stylized-ImageNet' to overcome the texture bias. By increasing the shape bias,~\cite{DBLP:conf/iclr/GeirhosRMBWB19} demonstrated improved performance and robustness. 
Since we use the ImageNet pre-trained ResNet-101 as a feature extractor, it is necessary to understand the nature of extracted features from the input images.}

{
Unlike `Stylized-ImageNet', in domain adaptation tasks, one cannot always expect to get such a curated dataset with ground truth labels on the target domain for each task. Instead, it might be easy and desirable to change the loss function that enhances the shape features with the same training dataset.
Surprisingly, in our observations, we find that the features learnt by the classifier indicate the high-level features or the saliency maps~\cite{DBLP:journals/corr/SimonyanVZ13} representing the shape of the object in the image. 
The contradistinguish loss is formulated and optimized in such a way that the features extracted are most unique and contrastive for a given image in comparison to other images in the dataset. This consequently is observed as the features corresponding to shapes in the form of silhouette in the feature visualizations in Fig.~\ref{fig:visda17 misclassification plots} as each object posses a unique shape as it's most contrasdistinguishing character, i.e., the character which is most discriminative and unique to the given image. 
}

\section{Concluding Remarks} \label{sec:conclusions and future work}
In this paper, we have proposed a direct approach to solve the problem of unsupervised domain adaptation that is different from the standard distribution alignment approaches. In our approach, we jointly learn a Contradistinguisher on the source and target domain distribution in the same input-label feature space using contradistinguish loss for unsupervised target domain to identify contrastive features. We have shown that contrastive learning overcomes the need and drawbacks of domain alignment, especially in tasks where domain shift is very high (e.g., language domains) and data augmentation techniques cannot be applied. Due to the inclusion of prior enforcing in the contradistinguish loss, the proposed unsupervised domain adaptation method \smallmethodname{} could incorporate any known target domain prior to overcoming the drawbacks of skewness in the target domain, thereby resulting in a skew-robust model. 
We validated the efficacy of \smallmethodname{} by experimenting on the synthetically created toy-dataset.
\color{black}\color{black}We further demonstrated the simplicity and effectiveness of our proposed method by performing multi-source domain adaptation on Office-31 and Digits datasets to consistently outperform other multi-source domain adaptation approaches.
\color{black}

{We have also tested the proposed method CUDA on the recent benchmark visual domain adaptation datasets such Office-31 and VisDA-2017 classification datasets and demonstrated above/on-par results with the state-of-the-art approaches.
We further analyzed the nature of the feature representation learnt using contradistinguish loss to identify the features related to the shapes that influence the predicted outcome. As the features related to shapes are learnt, we observed that it helps improving the performance and robustness of the trained model as the model is not biased to colors/textures in the images. We concluded that learning and improving shape bias is one of the keys to achieve ideal transfer learning and domain adaptation.
}

\section*{Acknowledgments}
The authors would like to thank the Ministry of Human Resource Development (MHRD), Government of India, for their generous funding towards this work through the UAY Project: {IISc 001}. 
The authors thank Tejas Duseja for helping the authors with setting up some experiments. 
The authors would also like to thank anonymous reviewers for providing their valuable feedback that helped in improving the manuscript.

\bibliographystyle{IEEEtran}
\bibliography{cuda_pami}

\begin{IEEEbiography}[{\includegraphics[width=1in,height=1.25in,clip,keepaspectratio]{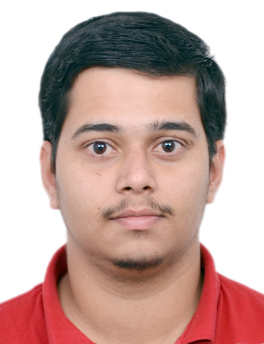}}]{Sourabh Balgi} is a doctoral candidate in causality and machine learning, advised by Prof. Jose M. Pe{\~n}a 
at the Statistics and Machine Learning (STIMA) division, Department of Computer and Information Science (IDA), Link{\"o}ping University, Sweden. 
He received his B.E. degree in Electronics and Communication Engineering from Sri Jayachamarajendra College of Engineering, Mysore, India in 2013. 
He worked at Mercedes-Benz Research and Development India, Bangalore as a software engineer from 2013 to 2016. He received his M.Tech. degree in Artificial Intelligence from Indian Institute of Science (IISc), Bangalore, India in 2019. 
He worked as a Research Associate at the Statistics and Machine Learning Lab, Department of Computer Science and Automation (CSA), Indian Institute of Science (IISc), Bangalore, India. 
His research interests include unsupervised deep learning for computer vision and machine learning with an emphasis on transfer learning, specifically domain adaptation and disentangled representation learning for domain adaptation and causal inference.
\end{IEEEbiography}

\begin{IEEEbiography}[{\includegraphics[width=1in,height=1.25in,clip,keepaspectratio]{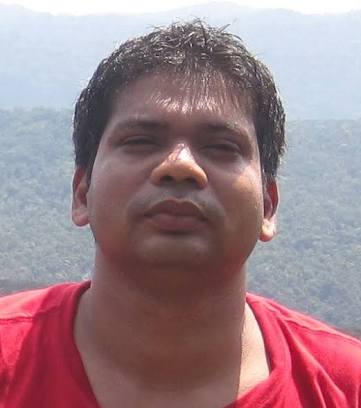}}]{Ambedkar Dukkipati} is an Associate Professor at the Department of Computer Science and Automation, IISc. He received his Ph.D. degree from the Department of Computer Science and Automation, Indian Institute of Science (IISc), Bangalore, India and B.Tech from IIT Madras. He held a post-doctoral position at EURANDOM, Netherlands. 
Currently, he also heads the Statistics and Machine Learning group at the Department of Computer Science and Automation, IISc. His research interests include statistical network analysis, network representation learning, spectral graph methods, machine learning in low data regime, sequential decision-making under uncertainty and deep reinforcement learning.
\end{IEEEbiography}

\balance







\end{document}